\documentclass[11pt, a4paper, singlecolumn, copyright, goog]{google}
\usepackage{smiles}
\usepackage[authoryear, sort&compress, round]{natbib}
\usepackage{wrapfig}
\usepackage{nicefrac}       % compact symbols for 1/2, etc.
\usepackage{microtype}      % microtypography
\usepackage{multicol}
\usepackage{transparent}
\usepackage{array}
\usepackage{bm}
\usepackage{caption}
\usepackage{tcolorbox}

\usepackage{xcolor, colortbl}  
\usepackage[dvipsnames]{xcolor}
\usepackage{listings}
\usepackage{minted}
\usepackage{pifont}  % for checkmark and xmark

\newcommand{\cmark}{\ding{51}} % check mark
\newcommand{\xmark}{\ding{55}}

\definecolor{darkblue}{rgb}{0, 0, 0.5}
\hypersetup{colorlinks=true, citecolor=darkblue, linkcolor=darkblue, urlcolor=darkblue}
\definecolor{corn}{rgb}{0.98, 0.93, 0.36}
\definecolor{atomictangerine}{rgb}{1.0, 0.6, 0.4}
\definecolor{ballblue}{rgb}{0.13, 0.67, 0.8}
\definecolor{green}{HTML}{009B55}
\definecolor{lightcoral}{HTML}{F08080}

\keywords{LLM-as-a-judge, Reinforcement Learning, Tool-Integrated Reasoning}
\uselogo{google}

\title{
Incentivizing Agentic Reasoning in LLM Judges via Tool-Integrated Reinforcement Learning}
\correspondingauthor{ranxu@google.com}

\reportnumber{0001} % Leave blank if n/a

\newcommand{\ours}{\texttt{TIR-Judge}}

\newcommand{\blue}[1]{{\textcolor{blue}{{{#1}}}}}
\newcommand{\red}[1]{{\textcolor{red}{{{#1}}}}}
\author[1,2*]{Ran Xu}
\author[2]{Jingjing Chen}
\author[2]{Jiayu Ye}
\author[2]{Yu Wu}
\author[3]{Jun Yan}
\author[1]{Carl Yang}
\author[2]{Hongkun Yu}

\affil[1]{Emory University}
\affil[2]{Google}
\affil[3]{Google Cloud AI Research}

\affil[*]{Work done during an internship at Google.}

\begin{abstract}
Large Language Models (LLMs) are widely used as judges to evaluate response quality, providing a scalable alternative to human evaluation. However, most LLM judges operate solely on intrinsic text-based reasoning, limiting their ability to verify complex constraints or perform accurate computation. Motivated by the success of tool-integrated reasoning (TIR) in numerous tasks, we propose \ours{}, an end-to-end RL framework for training LLM judges that integrates a code executor for precise evaluation. 
\ours{} is built on three principles: (i) diverse training across verifiable and non-verifiable domains, (ii) flexible judgment formats (pointwise, pairwise, listwise), and (iii) iterative RL that bootstraps directly from the initial model without distillation. 
On seven public benchmarks, \ours{} surpasses strong reasoning-based judges by up to 6.4\% (pointwise) and 7.7\% (pairwise), and achieves listwise performance comparable to Claude-Opus-4 despite having only 8B parameters. Remarkably, \ours{}-Zero -- trained entirely without distilled judge trajectories, matches the performance of distilled variants, demonstrating that tool-augmented judges can self-evolve through iterative reinforcement learning.
\end{abstract}

\begin{document}

\maketitle

\vspace{-0.5ex}
\section{Introduction}
\vspace{-0.5ex}
Large Language Model (LLM)-based judges are emerging as a critical component in the LLM ecosystem, typically used with scoring and ranking model outputs. 
This evaluation capability is essential at multiple stages of LLM development: during post-training, judges provide preference signals for  alignment~\citep{chen2025judgelrm,whitehouse2025j1}; at inference time, judges verify and select responses through best-of-N decoding~\citep{huang2025is}; and during evaluation, judges deliver reliable assessments without manual human assessment~\citep{li2024generative}. 
Thus, training accurate LLM-based judges is of great importance for building powerful language models.

\begin{wrapfigure}{r}{0.54\textwidth}
% \begin{figure}[t]
    \centering
    \vspace{-1ex}
    \includegraphics[width=0.96\linewidth]{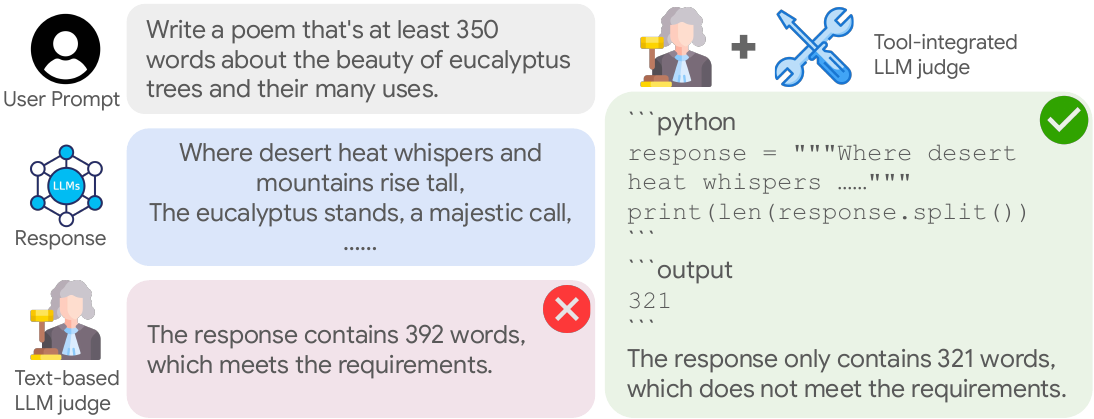}
    \vspace{-1ex}
    \caption{An example of LLM judge augmented with code execution, enabling precise judgments.}
    \label{fig:intro}
    \vspace{-1ex}
% \end{figure}
\end{wrapfigure}

Classical evaluation with reward models often outputs scores directly, which cannot fully harvest the inherent reasoning capability of LLM. Recent progress in generative reward modeling~\citep{genrm,zhao2025genprm} and reinforcement learning equips judges with thinking before producing final predictions~\citep{chen2025rm,whitehouse2025j1,guo2025reward,hong2025think}. 
While these approaches enhance judge quality by equipping LLMs with long chains of textual reasoning traces, they remain inherently limited in scenarios that require precise computation or symbolic reasoning -- capabilities that are much more challenging for text-only models~\citep{mirzadeh2025gsmsymbolic}.

% However, such text-only reasoning remains fundamentally limited in tasks demanding precise computation or symbolic manipulation~\citep{mirzadeh2025gsmsymbolic,zheng2025learning}.

% Training LLM-based judges beyond standard reward modeling is an active research frontier. Generative reward models (GRMs) address \emph{pointwise} evaluation by training LLMs to generate a chain-of-thought (CoT) explanation followed by a scalar reward or preference~\citep{genrm, ankner2024critique, yu-etal-2025-self}.
% More recently, advances in reinforcement learning from verifiable rewards (RLVR) have motivated the recast of  the judging task into a \emph{pairwise} setting, where binary comparison signals are optimized via RL to train \emph{pairwise} judges~\citep{chen2025rm,whitehouse2025j1,guo2025reward,hong2025think}. 
% While these approaches generally enhance judgment quality by equipping LLMs with long chains of textual reasoning traces, they remain inherently limited in scenarios that require precise computation or symbolic reasoning -- capabilities that are much more challenging for text-only models~\citep{mirzadeh2025gsmsymbolic,zheng2025learning}.
% \ran{} It remains a critical open challenge for developing judge models that can perform accurate calculations, verify symbolic logic, and produce reliable evaluations.

Recent advances in LLM tool-use provide a promising avenue to overcome the limitations of text-only judges~\citep{chen2023program,gao2023pal}. 
By granting access to executable interfaces for enumeration, verification, and computation, tools enable exact validation of reasoning steps rather than relying on potentially error-prone text-based inference. For example, code execution can automatically \emph{verify outputs on certain instructions}~\citep{zhou2023instruction} (as shown in Figure~\ref{fig:intro}) or \emph{check intermediate calculations} in math reasoning~\citep{xiong2025building}.
% By providing executable interfaces for enumeration, verification, and precise computation, tool-use enables exact validation of intermediate steps rather than relying solely on text-based reasoning. For instance, code can serve as an automatic verifier for instruction-following (IF) tasks~\citep{zhou2023instruction} or as a verifier to check intermediate calculations in math reasoning~\citep{lu2025mathcoder}. 
Early attempts have also explored equipping LLM judges with tool-use abilities~\citep{agentrm,findeis-etal-2025-external,li2024toolaugmented}, but these approaches reveal two major limitations. (i) \emph{Inference-time restriction}: most methods integrate tool-use only at the inference stage, preventing deeper integration between reasoning processes and tool execution. (ii) \emph{Narrow task coverage}: many are tailored to specific domains or specialized task types, which limits their applicability in general-purpose judging scenarios. 
% Consequently, existing approaches fall short of realizing the full potential of tool-augmented judges, where 
These gaps highlight the need for robust judges that  tightly couple reasoning with tool execution and be optimized end-to-end. 
% , a capability essential for building robust and broadly applicable tool-augmented judges.
% highlighting the need for models that can robustly  \ran{} code execution with reasoning across diverse and complex tasks.
% Early attempts have also explored equipping LLM-based judges with tool-use abilities~\citep{agentrm,findeis-etal-2025-external,li2024toolaugmented}, yet these efforts often remain constrained since (i) \emph{inference-time limitaion}: many of these works tool-use to the inference stage without fostering deeper interaction between reasoning processes and computational execution, (ii) \emph{narrow application scopes}: others target narrow domains or specialized task types~\citep{li2024toolaugmented}. Consequently, their applicability to broader judging scenarios is restricted, underscoring the need for judge models that can seamlessly and robustly integrate code execution with reasoning across diverse tasks.

Motivated by these challenges, our goal is to develop an LLM judge that can reliably integrate reasoning with code interpreter execution. 
Incorporating tool-integrated reasoning (TIR)~\citep{feng2025retool,li2025torl,xue2025simpletir,lin2025understanding}, we propose \ours{}, a framework that leverages reinforcement learning (RL) to teach models to generate code, execute it with interpreters, and iteratively refine their reasoning based on the resulting outputs. By reinforcing this cycle of reasoning and tool-use, \ours{} equips LLM judges with the ability at the training time to deliver more accurate and verifiable evaluations across diverse tasks.

Then, to fully unleash the potential of RL for  \ours{}, we introduce several key design choices. (i) \emph{Task diversity}: To balance between different tasks, we construct training prompts spanning both verifiable domains (e.g., competitive programming, mathematical reasoning) and non-verifiable domains (e.g., dialogue, safety, general coding), allowing the model to learn when tool invocation is beneficial and when pure reasoning suffices. (ii) \emph{Judgment flexibility}: To accommodate to different input/output formats, we diversify the evaluation tasks to cover pointwise, pairwise, and listwise ranking, ensuring broad applicability across practical use cases. (iii) \emph{Data efficiency}: 
% given the scarcity of annotated tool-integrated reasoning data especially in the judge domain, we design two training variants: 
unlike prior methods that rely on distillation as cold-start for RL~\citep{chen2025rm,hong2025think}, we demonstrate that \ours{} can bootstrap from the initial checkpoint. Specifically, \ours{}-Zero trains purely with iterative reinforcement learning for achieving self-improvement, while \ours{}-Distill provides an optional variant using a small amount of distillation data.
% \ours{}-Distill, which leverages a small amount of distillation data, and \ours{}-Zero, which relies purely on reinforcement learning without supervised data, enabling self-improvement through iterative evolution.

Our contribution can be summarized as follows:
\begin{itemize}[leftmargin=0.5cm]
\item We introduce \ours{}, a tool-integrated framework for training LLM-based judges with end-to-end multi-turn reinforcement learning. To the best of our knowledge, this is the first approach that jointly optimizes reasoning and tool-use for training LLM-based judges via RL.
\item We design several key strategies to fully exploit the power of reinforcement learning, including \emph{task diversification} across verifiable and non-verifiable domains, \emph{flexible judgment formats} (pointwise, pairwise, listwise), as well as an \emph{iterative RL scheme} that unlocks self-improvement in tool use even without distillation.
\item We evaluate \ours{} on seven public benchmarks covering diverse tasks and input formats. \ours{} consistently outperforms strong reasoning-based judges, achieving gains of up to 6.4\% (pointwise) and 7.7\% (pairwise). Moreover, \ours{} shows strong parameter efficiency: With only 8B parameters, it surpasses the 32B reasoning reward models on the PPE dataset, and reaches 96\% of the performance of Claude-Opus-4 in the listwise setting in RewardBench 2. 
Interestingly, \ours{}-Zero, the judge trained without any distillation, achieves a 1.2\% gain over its distilled counterpart at 4B scale, highlighting the power of RL to bootstrap reasoning and tool-use capabilities.
\end{itemize}

\section{Related Works}
\label{sec:related}
\textbf{Reasoning-Enhanced Reward and Judge Models.}
A growing line of work strengthens reward models (RMs) and judges by explicitly training them to \emph{reason} before issuing a score or decision.
Generative Verifiers~\citep{genrm} frame verification as next-token prediction with chain-of-thought, achieving notable gains on mathematical and algorithmic reasoning tasks. Other approaches strengthen judgment quality by generating critiques prior to reward prediction~\citep{ankner2024critique,yu-etal-2025-self,shen2024boosting}, leveraging multi-round preference optimization to progressively refine judge capability~\citep{wang2024self}, or teaching models to first propose evaluation plans or rubrics before producing a final judgment~\citep{evalplanner,openrubrics}. \citet{liu2025inference} further study how to allocate additional compute and structure critique signals to improve reliability. 
More recently, reinforcement learning has been used to optimize LLM-as-a-Judge~\citep{chen2025judgelrm,chen2025rm,whitehouse2025j1,guo2025reward,hong2025think}, encouraging longer, higher-quality reasoning and reducing bias across pairwise and pointwise settings~\citep{whitehouse2025j1}. While effective at strengthening textual reasoning and planning, these methods remain limited to reasoning expressed in natural language and often focus primarily on pairwise judgment.

\textbf{Tool-Assisted Reward and Judge Models.}
Another line of work augments judges with external tools. \citet{li2024toolaugmented} incorporate verifiable signals alongside preference data for judge training, though primarily within tool-use scenarios. 
\citet{agent-as-a-judge} evaluate agentic judge capabilities in agent settings, and Agentic Reward Modeling~\citep{agentrm} integrates human preferences with correctness checks to construct more reliable rewards. 
\citet{findeis-etal-2025-external} study whether external  tools (e.g., code execution, web search) improve LLM-as-a-Judge annotations, reporting consistent but task-dependent gains. 
However, these approaches largely depend on \emph{prompted} tool use rather than training judges to \emph{learn when and how} to invoke tools and to integrate their outputs into decisions.

\textbf{Reinforcement Learning for Tool-integrated Reasoning.}
Recent work explores reinforcement learning for TIR. \citet{feng2025retool,bai2025towards,li2025torl} train LLMs to interleave reasoning with code execution, discovering strategic tool-use policies that improve math and programming tasks. 
Several works extend this paradigm by interleaving reasoning steps on search agents~\citep{jin2025searchr,song2025r1,acesearcher}, web agents~\citep{qi2025webrl,zhuang2025workforceagent} and coding agents~\citep{du2025generalizable}. Other studies focus on optimal reward design for TIR~\citep{dong2025tool,wang2025acting} or provide theoretical analysis of its advantages~\citep{lin2025understanding}.
% ToolRL argues that reward alone suffices for learning tool use, dispensing with specialized supervision.
% OTC formulates \emph{optimal tool calls} to encourage efficient, minimal actions through RL \cite{wang2025otc}.
% These methods target task-solving agents rather than judges. Our work differs by \emph{training a judge} to use tools for \emph{verification}, combining SFT on filtered tool-augmented trajectories with RL to learn a policy over tool decisions and the integration of executable feedback.
\vspace{-0.5ex}
\section{Preliminaries}
\vspace{-0.5ex}
\textbf{Problem Setup.}
\label{sec:problem_setup}
% \vspace{-0.5ex}
We consider the task of \textit{LLM-based judgment}: given a user prompt $x \in \mathcal{X}$ and $n$ model-generated responses $\mathcal{Y} = \{y_1, y_2, \ldots, y_n\}$, the goal is to evaluate the quality of responses for the prompt. 
The judge model $J_\theta$ produces an evaluation output conditioned on $(x, \mathcal{Y})$. In this work, we consider three evaluation settings: (i) \textbf{Pointwise evaluation}: given $(x, y)$, the judge assigns a scalar score, $J_\theta(x, y) = s_\theta(x, y) \in \mathbb{R}$; 
(ii) \textbf{Pairwise evaluation}: given $(x, y_a, y_b)$, the judge selects the preferred response,
$J_\theta(x, y_a, y_b) = \arg \max_{i \in \{a,b\}} s_\theta(x, y_i),$ where $s_\theta$ denotes a learned scoring function. This is also the most common evaluation setting; 
(iii) \textbf{Listwise evaluation}: given $(x, \mathcal{Y})$ with $n > 2$, the judge returns the index of the best response, $J_\theta(x, \mathcal{Y}) = \arg \max_{i \in \{1, \ldots, n\}} s_\theta(x, y_i).$ These settings unify a broad range of evaluations under a common framework\footnote{Note that in our work, the reference answer is \emph{unseen} during evaluation, different from the \emph{verification setting} \citep{li2025verifybench,yan2025verifybench} where the {reference answer} is also a part of the input.}.

\textbf{Tool-Augmented Judge.}
% \vspace{-0.5ex}
We extend the judge with the ability to call an external \textit{Python execution environment} $\cI$.    
For the prompt $x\in\cX$, At step $k$, the judgment trajectory $s_k$ is represented as \( s_k = \{r_1, c_1, o_1, \ldots, r_k, c_k, o_k \} \),
where $r_i$ is a natural language reasoning step, $c_i$ is a generated code, and $o_i = I(c_i)$ is the execution result of $c_i$~\citep{li2025torl}. The iterative process is defined as:
\begin{align}
(r_k, c_k) \sim J(x \oplus s_{k-1}), \quad 
o_k = \cI(c_k), \quad
s_k = s_{k-1} \oplus r_k \oplus c_k \oplus o_k.
\end{align}
This cycle continues until the judge produces a final prediction $a_i \sim J(x \oplus s_T)$ with $T$ being the final step. Unlike traditional text-only reasoning, the trajectory now interleaves reasoning, code execution, and tool feedback, enabling the judge to ground its decision in verifiable evidence.

\vspace{-0.5ex}
\section{Training \ours{}}
\begin{figure}[t]
    \centering
    % \vspace{-1ex}
    \includegraphics[width=0.99\linewidth]{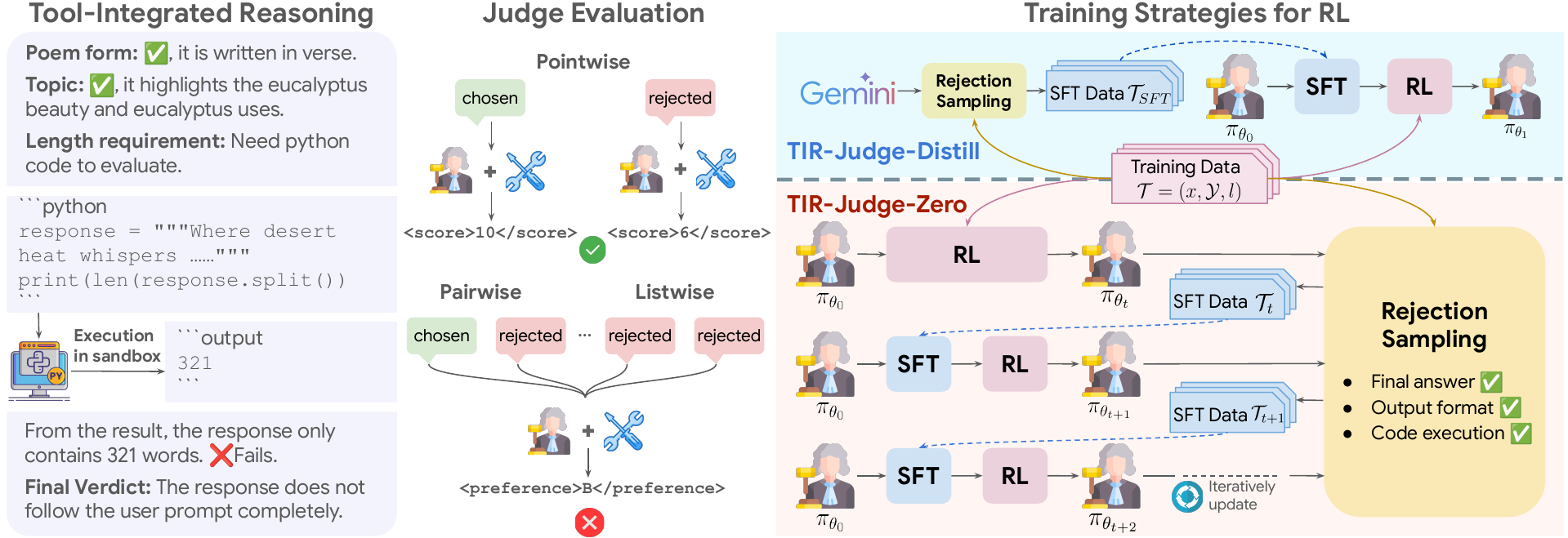}
    \caption{Overall framework of \ours{} variants. \ours{} natively supports tool use during judgment and is designed to handle diverse input formats.}
    \label{fig:framework}
\end{figure}

\vspace{-0.5ex}
We now describe the training procedure for \ours{}, which consists of four components: (1) data collection and filtering for RL, (2) the RL framework for training judges with integrated code execution tools, (3) reward design for RL, and (4) cold-start and iterative training strategies in RL. The overall framework of \ours{} is exhibited in Figure \ref{fig:framework}.

\subsection{Data Collection and Filtering}
\vspace{-0.5ex}
\label{sec:collect}
High-quality training data is crucial for RL with tool-augmented judges. 
Since judgment requires both prompts and candidate responses, we curate a collection of $(\text{prompt}, \text{responses})$ tuples spanning multiple tasks. 
Our corpus integrates both human-annotated preference data and automatically generated synthetic pairs to ensure diversity and scalability.

\textbf{Real Preference Pairs.}  
We sample human-labeled preference pairs from a variety of domains:  
\textbf{general helpfulness} --- \emph{HelpSteer 3}~\citep{wang2025helpsteer3};  
\textbf{reasoning} --- \emph{UltraInteract}~\citep{yuan2025advancing}, \emph{S1}~\citep{muennighoff2025s1};  
\textbf{coding} --- \emph{CodeRM}~\citep{ma2025dynamic};  
\textbf{instruction following (IF)} --- preference pairs from \emph{Tulu 3}~\citep{lambert2024tulu};  
\textbf{safety} --- \emph{Safe-RLHF}~\citep{dai2024safe}.  
Each prompt is paired with one preferred (\emph{chosen}) response and one or more  \emph{rejected} responses.

\textbf{Synthetic Preference Pairs.}  
Because reasoning preference data is often limited in scale, we augment the corpus with \emph{synthetic} preference pairs generated from verifiable prompts. 
For each prompt, we sample responses from multiple open-source models, including \emph{Qwen3-8B/14B}~\citep{qwen3technicalreport}, \emph{Gemma-2-9B}~\citep{team2024gemma}, and \emph{Gemma-3-12B}~\citep{team2025gemma}. 
The responses are automatically evaluated against verifiable functions (for IF tasks) or ground-truth solutions (for reasoning tasks) to form preference pairs.  
For \textbf{IF}, we use verifiable prompts from \emph{Tulu-3}~\citep{lambert2024tulu}, where correctness can be programmatically verified using lexical or structural constraints.
For \textbf{reasoning}, we employ  \emph{MATH}~\citep{MATH} and \emph{DAPO-Math}~\citep{yu2025dapo} for math domain and \emph{WebInstruct}~\citep{ma2025general}, and \emph{Loong}~\citep{huang2025loong} for general domain, both of which provide ground-truth solutions for exact verification.

In total, our dataset comprises approximately 26k preference pairs, including pointwise, pairwise, and listwise annotations, covering diverse domains such as helpfulness, reasoning, coding, safety, and verifiable instruction following.  
We apply strict 8-gram decontamination to eliminate any overlap between training prompts and evaluation benchmarks~\citep{oren2024proving}. 
This diverse mixture of data provides a strong foundation for training robust tool-augmented judges.

\subsection{Tool-Integrated RL with Verifiable Rewards}
\label{sec:reward}
\vspace{-0.5ex}
\textbf{Overall Framework.}
We adopt DAPO~\citep{yu2025dapo}, an improved variant of GRPO~\citep{shao2024deepseekmath}, for training the LLM judge $J$ parameterized by $\pi_{\theta}$.  
Given a prompt--answer pair $(q,a)$, we first sample a group of $G$ rollouts $\{s_i\}_{i=1}^G$ from the current policy $\pi_{\theta_{\text{old}}}$.  
Each rollout $s_i$ is assigned a scalar reward $R_i = R(s_i, a)$ with access to the oracle answer $a$.   
The policy $\pi_{\theta}$ is then updated with the following clipped policy gradient objective:
% \begin{small}
% \begin{align}
% \setlength{\abovedisplayskip}{3pt}
% \setlength{\belowdisplayskip}{3pt}
% \nonumber
% & \mathcal{J}(\theta)=\mathbb{E}_{(q, a) \sim \mathcal{D},\left\{s_i\right\}_{i=1}^G \sim \pi_{\theta_{\mathrm{old}}}(\cdot \mid q)} \\
% \nonumber
% & \quad \quad\left[\frac{1}{\sum_{i=1}^G\left|s_i\right|} \sum_{i=1}^G \sum_{t=1}^{\left|s_i\right|}\left(\min \left(r_{i, t}(\theta) \hat{A}_{i, t}, \operatorname{clip}\left(r_{i, t}(\theta), 1-\varepsilon, 1+\varepsilon\right) \hat{A}_{i, t}\right)-\beta D_{\mathrm{KL}}\left(\pi_\theta| | \pi_{\mathrm{ref}}\right)\right)\right] \\
% \nonumber
% & \quad \text{ s.t. } \quad 0<\mid\left\{s_i \mid \texttt{is\_equivalent}\left(a, s_i\right)\right\} \mid<G,
% \end{align}
% \end{small}

\setlength{\abovedisplayskip}{-4pt}
\setlength{\belowdisplayskip}{-2pt}
\begin{multline}
\nonumber
\mathcal{J}(\theta) = \mathbb{E}_{(q, a) \sim \mathcal{D},\{s_i\}_{i=1}^G \sim \pi_{\theta_{\mathrm{old}}}(\cdot \mid q)} \bigg[
\frac{1}{\sum_{i=1}^G |s_i|} \sum_{i=1}^G \sum_{t=1}^{|s_i|}
 \big( \min(r_{i, t}(\theta) \hat{A}_{i, t}, \\
 \operatorname{clip}(r_{i, t}(\theta), 1-\varepsilon_{\text{low}}, 1+\varepsilon_{\text{high}})\hat{A}_{i, t}) 
- \beta D_{\mathrm{KL}}(\pi_\theta \| \pi_{\mathrm{ref}})\big)\bigg]
\quad \text{s.t. } 0<|\{s_i : \texttt{is\_equivalent}(a, s_i)\}|<G
\end{multline}
% \end{small}

where $r_{i, t}(\theta)=\frac{\pi_\theta\left(s_{i, t} \mid q, s_{i,<t}\right)}{\pi_{\theta_{\text {old }}}\left(s_{i, t} \mid q, s_{i,<t}\right)}$ is the token-level weight, $\hat{A}_{i, t}=\frac{R_i-\operatorname{mean}\left(\left\{R_i\right\}_{i=1}^G\right)}{\operatorname{std}\left(\left\{R_i\right\}_{i=1}^G\right)}$ is the advantage at the token level, and \texttt{is\_equivalent} step filters out the prompts with accuracy equal to 1 and 0.
The hyperparameters $\varepsilon_{\text{low}}$ and $\varepsilon_{\text{high}}$ control the clipping range for importance weights, while $\beta$ regulates the KL divergence penalty to stabilize training.

\textbf{Additional Designs.}  
Beyond standard RL training, we implement two enhancements to stabilize tool-augmented judgment:
\emph{(i) Error Message Processing.}   
We truncate the outputs from Interpreter $\cI$ to only the final error line to avoid excessive context length while preserving useful feedback in $s_k$; 
\emph{(ii) Sandbox Output Masking.}   
Since execution results $o_i = \cI(c_i)$ may cause the model to overfit by memorizing outputs, we mask $o_i$ during loss computation, following~\citet{li2025torl,jin2025searchr}. This prevents reliance on exact strings and improves training stability.  

\textbf{Reward Designs.} 
To effectively facilitate multi-turn RL with code execution, we design a structured covering three aspects, described as follows:

(i) \emph{Correctness Reward} $R_{c}$: 
This component measures whether the judge’s prediction aligns with the reference preference label.  
Let $x$ denote the prompt, $\cY = \{y_1, \ldots, y_n\}$ the candidate responses, and $l$ the ground-truth preferred response.  
The reward is defined as:
\begin{equation}
\setlength{\abovedisplayskip}{10pt}
\setlength{\belowdisplayskip}{10pt}
R_c =
\begin{cases}
\mathbb{I}\big(s_\theta(x, y_{\text{pos}}) > s_\theta(x, y_{\text{neg}})\big), & \text{for pointwise evaluation}, \\[6pt]
\mathbb{I}\big(J_\theta(x, \cY) = l\big), & \text{for pairwise or listwise evaluation}, \\[6pt]
0, & \text{otherwise},
\end{cases}
\end{equation}
where $\mathbb{I}(\cdot)$ is the indicator function, $s_\theta(x,y)$ denotes the judge’s scoring function, and $J_\theta(x,\cY)$ is the predicted best response under the judge’s policy.   
Intuitively, $R_c = 1$ if the judge’s decision matches the ground-truth preference, and $R_c = 0$ otherwise.

(ii) \emph{Format Reward} $R_f$: 
To ensure reliability, the judge is required to strictly follow a predefined structured output format.  
Specifically, prediction scores must be enclosed within \texttt{<score>} and \texttt{</score>} tags, the preference label must be wrapped in \texttt{<preference>} and \texttt{</preference>} tags, and all code segments must be enclosed using \verb|```python| and \verb|```|.  
In addition, to accommodate both \emph{reasoning} and \emph{non-reasoning} tasks and \emph{discourage unnecessary tool calls}, we introduce a heuristic: for safety and general helpfulness prompts, a positive format reward is granted only if the model produces a valid output \emph{without} invoking tools.  
Formally, $R_f = 1$ if the output satisfies all formatting constraints (and the no-tool heuristic when applicable), and $R_f = 0$ otherwise.

(iii) \emph{Tool-Specific Reward} $R_t$: We encourage accurate and efficient tool use by penalizing errors or excessive executions \citep{wang2025acting}. We set the max number of tool calls per trajectory to 3, and set $R_t=1$ only when code blocks $c_i$ are error-free and within the call budget; otherwise $R_t=0$.   

The final reward $R$ is defined as a combination of correctness, format, and tool-specific rewards and assigns full credit only when correctness, format, and tool-use are all satisfied:
% \begin{equation}
% \setlength{\abovedisplayskip}{5pt}
% \setlength{\belowdisplayskip}{5pt}
% R =
% \begin{cases}
% 1, & \text{if } R_t = 1 \;\wedge\; R_f = 1 \;\wedge\; R_c = 1, \\[6pt]
% 0.1, & \text{if } R_c = 1 \text{ but } (R_t=0 \;\vee\; R_f=0), \\[6pt]
% 0, & \text{if } R_c = 0.
% \end{cases}
% \end{equation}
\begin{equation}
\setlength{\abovedisplayskip}{5pt}
\setlength{\belowdisplayskip}{5pt}
R \;=\; R_c \times \bigl(0.1 + 0.9\,\mathbb{I}[R_t=1 \land R_f=1]\bigr).
\end{equation}
% The final reward $R$ , partial credit when only correctness holds, and zero otherwise.% This encourages the judge to generate both high correct answers and high-quality tool calls.
\vspace{-2ex}
\subsection{Training Strategies for RL}
% \vspace{-1ex}
Directly applying RL often leads to suboptimal outcomes, as the base model lacks sufficient reasoning and tool-use capability.  
To address this, we design two cold-start strategies for training \ours{}.  

\noindent \textbf{Distillation from Teacher Models (\ours{}-Distill).}  
We leverage a stronger teacher, \texttt{Gemini-2.5}-\texttt{Flash} with code execution~\citep{comanici2025gemini}, to generate high-quality trajectories via rejection sampling.  
For each user prompt $x$ and corresponding $\mathcal{Y}$, we collect a trajectory $s$ and a final prediction $a$ as $(x, \mathcal{Y}, s, a) \sim J$.  
Only trajectories that produce correct answers are retained, yielding a dataset $\cT_{\text{SFT}} = \{(x, \mathcal{Y}, s, a) \mid R(s, a, l) = 1\}$. 
Then the student judge is trained via supervised fine-tuning (SFT) with objective  
\begin{equation}
\setlength{\abovedisplayskip}{5pt}
\setlength{\belowdisplayskip}{5pt}
\mathcal{L}_{\text{SFT}}
= - \mathbb{E}_{(x, \tau) \sim \mathcal{T}_{\text{SFT}}}
  \left[\sum_{i=1}^{|y|} \log f_\theta(\tau_i \mid \tau_{<i}, x)\right],
\end{equation}
where $\tau=(s, a)$ is the target trajectory with reasoning and code steps.  
As in RL training,  \emph{interpreter feedback tokens are masked} to prevent learning on execution results. 
In total, we collect about 10k tool-integrated trajectories for SFT, which serve as the initialization before reinforcement learning.

\noindent \textbf{Iterative Training without Distillation (\ours{}-Zero).}  
Beyond teacher distillation, we investigate whether tool-augmented judges can improve purely through \emph{self-bootstrapping}~\citep{yuan2024selfrewarding,huang-etal-2023-large,zelikman2022star,xiong2025building,singh2024beyond}.  
The process alternates between RL, rejection sampling, and supervised fine-tuning.  

Starting from the initial model $\pi_{\theta_0}$, we first obtain the checkpoint $\pi_{\theta_1}$ via direct RL on training data  as $\pi_{\theta_1} \leftarrow \mathrm{RL}(\pi_{\theta_{0}})$ (Sec.~\ref{sec:reward}).  
Then, for each prompt $x$, we sample multiple trajectories from $\pi_{\theta_1}$ as $\{s_i\}_{i=1}^G \sim \pi_{\theta_t}(\cdot \mid x)$ ($G=4$ in our study),  where each trajectory contains reasoning, code, and execution results:  
$s_i = \{r_1, c_1, o_1, \ldots, r_k, c_k, o_k\}$.  
We retain only valid trajectories that (i) produce the correct answer $l$, (ii) satisfy the output format, and (iii) execute without interpreter errors as 
$\mathcal{T}_t = \{(x, s, a) \mid R(s, a, l) = 1\}$.  
To promote efficiency, for each prompt we further keep only one trajectory, preferring the shortest response or the one with the fewest tool calls.   
The dataset $\mathcal{T}_t$ is then used for SFT, and the fine-tuned model initializes the next RL round.  
After each cycle, we select the best checkpoint based on held-out validation accuracy and repeat the RS $\to$ SFT $\to$ RL loop: 
% Formally, the iterative dynamics are:  
\begin{equation}
\setlength{\abovedisplayskip}{5pt}
\setlength{\belowdisplayskip}{5pt}
\nonumber
\mathcal{T}_{t+1} \leftarrow \mathrm{RS}(\pi_{\theta_t}), 
\quad
\pi_{\theta_{t+1}} \leftarrow \mathrm{SFT}(\pi_{\theta_0}, \mathcal{T}_{t+1}), 
\quad
\pi_{\theta_{t+1}} \leftarrow \mathrm{RL}(\pi_{\theta_{t+1}}).
\end{equation}
This iterative process allows \ours{}-Zero to progressively bootstrap stronger reasoning and tool-use capabilities entirely from a initial model and facilitates self-improvement without distillation.

\vspace{-1ex}
\section{Experiments}
\vspace{-1ex}
\label{sec:exp}
\subsection{Experiment Setups}
\label{sec:exp_setup}
\textbf{Evaluation Datasets.} 
Following prior work~\citep{whitehouse2025j1,chen2025rm}, we focus  on \emph{reasoning tasks}, evaluating \ours{} on PPE Correctness~\citep{ppe}.  
We additionally consider two more challenging datasets on judges: IFBench~\citep{agentrm} for instruction-following and CodeJudgeBench~\citep{codejudgebench} for code generation.  
All evaluations are conducted under both \emph{pointwise} and \emph{pairwise} settings to demonstrate the broader applicability of \ours{}.  
We also evaluate on \emph{general-domain} judge benchmarks, where reasoning constitutes a subset, including RewardBench~\citep{rewardbench}, RM-Bench~\citep{liu2025rmbench} and JudgeBench \citep{tan2025judgebench} for pointwise/pairwise evaluation, and RewardBench 2~\citep{rewardbench2} for \emph{listwise} evaluation.  
 
\textbf{Implementation Details.}  
We use \texttt{Qwen3-8B} and \texttt{Qwen3-4B-Instruct-2507}~\citep{qwen3technicalreport} as backbones, without enabling thinking mode, and implement training with Verl-Tool~\citep{jiang2025verltool}.   
For SFT, we train with batch size 64, learning rate 2e-6, context length 8192, for 1 epoch.  
For RL, we set the micro batchsize per gpu to 4, mini batchsize to 128 and number of rollout to 8. We set $\varepsilon_{\text{low}}=0.2, \varepsilon_{\text{high}}=0.3, \beta=0.01$, max response length to 8192, learning rate 1e-6 and train for 2 epochs. 5\% of the prompts from each task were hold-out for validation. The 
experiments are run with 8 NVIDIA H100 80G GPUs. 
For data collection in Sec. \ref{sec:collect}, we generate 2 rollouts for each model with $t=0.9, p=0.95$. No external feedback (e.g., GPT annotations) is used.
\begin{table*}[t]
\centering
\vspace{-1ex}
\renewcommand\arraystretch{0.95}
\caption{Main results on six benchmarks. $^\dagger$ indicates results reported from the original papers. 
CJBench, RWBench, and JGBench denote CodeJudgeBench, RewardBench, and JudgeBench. 
``Distill?'' specifies whether the model relies on additional judge data distilled from teacher models. 
\textbf{Bold} highlights the overall best accuracy, while \blue{blue} and \red{red} mark the best results within our direct comparisons for \blue{pointwise} and \red{pairwise} settings, respectively. \vspace{-1ex}}
% \caption{Main Experiments on six benchmarks. $^\dagger$: Results copied from corresponding papers. CJBench, RWBench, JGBench stands for CodeJudgeBench, RewardBench and JudgeBench. ``Distill?'' specifies whether the model requires additional judge data from teacher models. \textbf{bold} specified the overall best accuracy, and \blue{blue}m \red{red} standas for best accuracy within our direct comparison for pointwise and pairwise, respectly.}
\label{tab:main_ppe}
\renewcommand\arraystretch{0.95}
\resizebox{1.01\linewidth}{!}{ %
\begin{tabular}{l | cc | c  c c c c c| c c | c c c}
\toprule
\multirow{2.5}{*}{\textbf{Baselines}} & \multirow{2.5}{*}{\textbf{|Train|}} & \multirow{2.5}{*}{\textbf{Distill?}} & \multicolumn{6}{c|}{\textbf{PPE Correctness}} &  \multirow{2.5}{*}{\textbf{IFBench}} & \multirow{2.5}{*}{\textbf{CJBench}} & \multirow{2.5}{*}{\textbf{RWBench}} & \multirow{2.5}{*}{\textbf{RMBench}}& \multirow{2.5}{*}{\textbf{JGBench}}
\\
\cmidrule(lr){4-9}
& & &  MMLU-P & MATH & GPQA & MBPP-P & IFEval & Avg.  & & &  \\
\midrule
\multicolumn{10}{l}{\bf LLM-as-a-Judge (Pairwise Evaluation unless specified)} \\
\midrule
Qwen3-4B-Instruct (Pointwise) & -- &  -- &64.3 &	83.1	 &38.0 &	62.4 &	55.2 &	60.6 &	56.2 &	16.6 & 76.5	& 66.9	 & 50.8\\ 
Qwen3-8B (Pointwise) &  -- & -- & 68.7 &	64.2 &	56.5	 &58.9 &	57.4 &	61.1 &	55.9 &	54.9 &	79.2 &	69.3 &	64.9 \\ 
Gemini-2.5-Flash (Pointwise) &  -- & -- &56.5 &	79.5	 &46.4 &	63.0 &	63.9 &	61.9 &	51.6 & 53.3 & 80.7	& 70.8 & 66.9 \\ \midrule
GPT-4o$^\dagger$ & -- & -- & -- & -- & -- & -- & -- & 57.6 & 61.3 & -- & 86.7	& 72.5 & 56.6 \\
GPT-o1-mini$^\dagger$  & -- & -- &  -- & -- &-- & -- &--  & 71.3  & 70.1 & -- & 87.1& -- & 65.7\\
DeepSeek-R1-671B$^\dagger$ &  -- & -- & -- & -- & -- & -- & -- & \textbf{76.5} & 68.0 & -- & 90.6 & -- & 73.1 \\
% Llama-3.3-70B-Instruct$^\dagger$ & -- & -- & 72.1 & 73.1 & 61.2 & 59.6 & 62.3 & 65.7 & -- & -- \\
Claude 3.5$^\dagger$ & --&-- & 81.0 & 86.0 & 63.0 & 54.0 & 58.0 & 68.4 & -- & 58.3 & 84.2	& 61.0 & 64.3 \\ 
Qwen3-4B-Instruct (Pairwise) &  -- & -- & 63.9	 &83.1	 &35.0	 &59.7 &	60.7 &	60.4 &	62.2	 &34.5 & 86.0 & 	 75.3	 & 63.9\\ 
Qwen3-8B (Pairwise) &  -- & -- & 73.8 &	80.2 &	57.3 &	57.8 &	58.4 &	65.5 &	61.3 &	60.8 & 87.0	 & 77.9	 & 67.5 \\ 
Gemini-2.5-Flash (Pairwise) &  -- & -- & 68.8	&85.5	 &58.1 &	86.5 &	75.0 &	74.8 &	69.3 & 66.5 & 93.4	& 81.9	& 75.4\\ 
\midrule
\multicolumn{10}{l}{\bf Scalar Reward Models (Pointwise)} \\
\midrule
Armo-RM-8B$^\dagger$ & 1000k & \ding{55} & 66.0 & {71.0} & 57.0 & 54.0 & 58.0 & 61.2 & 62.9 &  -- & 90.3	 & 67.7 &--  \\
Skywork-Gemma-2-27B$^\dagger$ &  80k & \ding{55} & 55.0 & 46.2 & 44.7 & 69.1 & 58.3 & 54.7 & 63.2 & -- & \textbf{93.8} & 67.3 & --  \\
Deepseek-BTRM-27B$^\dagger$ & 237k &  \ding{55} & 68.8 & 73.2 & 56.8 & 68.8 & 66.0 & 66.7 & -- & -- & 81.7 & -- & --  \\
% Skywork-RM-v2 & 40M & & & & & & & 87.2 \\
\midrule
\multicolumn{10}{l}{\bf Text-based Reasoning Judges (Pairwise Evaluation unless specified)} \\
\midrule
Deepseek-GRM-27B$^\dagger$ & 237k  & \ding{55} & 64.8 & 68.8 & 55.6 & 50.1 & 59.8 & 59.8 & -- & -- & 86.1 & -- & -- \\	
% Evalplanner-8B$^\dagger$ & 22k & \ding{55} & 57.0 & 59.0 & 50.3 & 47.7 & 50.0 & 52.8 & -- & -- & 83.1 &	68.1 &	30.2 \\
J1-8B (Pairwise)$^\dagger$ & 22k& \ding{55}& 65.6 & 70.0 & 53.2 & 53.1 & 54.0 & 59.2 & -- & -- & 85.7 &	73.4	 &42.0\\
J1-8B (Pointwise)$^\dagger$ & 22k & \ding{55}  & -- & -- &  -- & -- &--  & 58.5  & -- & -- & -- & -- & -- \\
RRM-7B & 420k &  \ding{55}& 66.5 & 88.0 & 57.9 & 61.2 & 53.6 & 65.4 & 60.1 & 63.4 &82.2 &	70.4 &  67.0 \\
RM-R1-Deepseek-Distill-7B & 73k & \ding{51} & 67.3 & 91.2 & 62.6 & 60.5 & 53.0 & 66.9 & 56.6 & 63.2 & 80.1	 &72.4 & 67.7 \\
RM-R1-Instruct-7B & 73k & \ding{51} & 64.1 & 74.5 & 60.7 & 57.3 & 57.8 & 62.9 & 59.0 & 57.5 &85.2 &	70.2 & 60.3 \\
Think-RM 7B & 10k & \ding{51} & 66.5 & 78.3 & 55.6 & 58.1 & 63.9 & 64.5 & 57.4 & 54.6 & 86.0 &	73.9 & 64.6 \\
\midrule 
\multicolumn{10}{l}{\bf Tool-augmented Judges} \\
\midrule
Qwen3-4B-Tool (Pointwise) & -- & -- & 64.6 &	81.6 &	38.3 &	61.0 &	49.8 &	59.1 &	44.1 &	18.0 & 78.4 &	72.1 &	56.6\\		
Qwen3-8B-Tool (Pointwise) & -- & -- & 67.0	&72.4	&54.0&	56.0&	34.0	&56.7&	27.1	&45.9& 78.0 &	67.9 &	59.4\\	
Gemini-2.5-Flash-Tool (Pointwise) & -- & -- & 68.2	 & 86.0 &	48.9 &	58.7 &	73.5 &	67.1 &	53.0  & 47.9 & 81.3 & 71.2  & 66.5\\
\rowcolor{magenta!15}  \ours{}-Distill 4B (Pointwise) &  26k & \ding{51} & 58.7 &	81.9 &	45.8	 &64.1 &	78.9	 &65.9	 &65.8 &	59.9 & 76.6 &	71.9 &	66.7 \\ 
\rowcolor{magenta!15}  \ours{}-Zero 4B (Pointwise) & 26k & \ding{55}&  62.5 &	87.3 &	54.7 &	64.8 &	79.8 &	69.8	 &65.9 &	  61.5 & 77.3	 & 72.8 & 	70.4\\ 
\rowcolor{teal!15}  \ours{}-Distill 8B (Pointwise)& 26k & \ding{51} & 70.9 &	88.1 &	52.3 &	61.0	 &83.0	 & \blue{71.0}	 & \blue{68.4} &	\blue{61.9} & 81.0	& 76.7	 & \blue{68.2}\\
\rowcolor{teal!15}  \ours{}-Zero 8B (Pointwise) & 26k & \ding{55} & 67.8 &	88.0	 &53.2 &	64.7 &	77.8 &	70.3	 & 66.8 &	60.8 & \blue{81.4}	 & \blue{76.3}	 & 67.5  \\  \midrule
AgentRM 8B + 8B (Pairwise) & -- & -- & 64.6 & 76.0 & 52.8 & 61.7 & 73.0 & 65.6 & 67.0 & 59.2 & 87.7 & 69.7 & 59.4  \\
Qwen3-4B-Tool (Pairwise) & -- & -- & 63.5	& 83.3 &	35.9 &	58.9 &	62.3 &	60.8 &	59.2 &	29.2 & 85.2 & 	75.7 & 	63.0\\							
Qwen3-8B-Tool (Pairwise) & -- & -- & 72.0 &	85.2 &	56.0 &	54.3 &	60.8 &	65.7 &	52.5	 &54.9	 & 86.2	 &77.3 &	65.9\\	
Gemini-2.5-Flash-Tool (Pairwise) & -- & -- & 73.1 &	87.5 &	60.2 &	85.2 &	84.0	 &78.0 &	68.5 & 66.3 & 90.1 & 80.9 & 74.6 \\	

\rowcolor{magenta!15}  \ours{}-Distill 4B (Pairwise) & 26k & \ding{51} & 69.0 &	88.7 &	54.8 &	60.6 &	83.6 &	71.3	 &73.7 &	69.8 &   	 87.7 & 	78.0	&  70.5\\ 
\rowcolor{magenta!15}  \ours{}-Zero 4B (Pairwise)& 26k & \ding{55}  &  75.0	 & 93.3 &	61.7 &	67.3	 &84.5	 & \red{76.3}	 &70.3	 &\red{70.8} & 86.7 &	80.8 &	\red{73.7} \\ 
\rowcolor{teal!15}  \ours{}-Distill 8B (Pairwise)& 26k & \ding{51} & 72.2	& 90.4	 &53.8	 &63.2	 &85.7	 &73.0 &	\red{\textbf{74.3}} &	70.0 &	87.9 &	82.2 &	72.6 \\
\rowcolor{teal!15}  \ours{}-Zero 8B (Pairwise) & 26k & \ding{55} & 76.6	&94.0&	58.5	&68.8&	80.8	&\underline{75.7}	&68.9	&69.3	&\red{89.1}&	\red{83.7}	&72.0  \\
\midrule
\multicolumn{10}{l}{\bf For Reference: Text-based Reasoning Judge Baselines with >10B Parameters (Pairwise Evaluation)} \\
\midrule
% Evalplanner 70B$^\dagger$ & 22k & \ding{55}  & 78.4 & 81.7 & 64.4 & 62.2 & 64.3 & 70.2 & -- & -- \\
J1 70B$^\dagger$ & 22k & \ding{55} & 79.0 & 86.0 & 65.9 & 66.0 & 67.3 & 72.8 & -- & -- & 93.3 & 	82.7 & 60.0\\
RRM 32B & 420k &  \ding{55} & 80.5 & 94.3 & 68.4 & 72.8 & 60.2 & 75.3 & 60.8 & \textbf{76.3} & 91.2 & \textbf{85.4} & 76.0 \\
RM-R1-Deepseek-Distill-14B & 73k & \ding{51} & 78.8 & 94.5 & 63.3 & 70.5 & 63.0 & 74.0 & 58.6 & 65.5 &88.9 &	81.5   & 76.2 \\
% RM-R1-Instruct-14B & 73k & \ding{51}& 68.6 & 77.6 & 55.4 & 60.6 & 58.3 & 64.1 & 58.2 & 63.2 & 88.3 &	78.7 \\
RM-R1-Deepseek-Distill-32B& 73k & \ding{51} & 79.8 & 95.4 & 65.2 & 74.6 & 63.3 & 75.6 & 60.4 & 65.8 & 90.9 &	83.9  & \textbf{78.4}\\
% RM-R1-Instruct-32B & 73k & \ding{51} & 71.0 & 81.3 & 57.2 & 64.8 & 60.4 & 66.9 & 60.4 & 63.7 & 91.4 &	79.1 \\
\bottomrule
\end{tabular}%
}
\vspace{-1ex}
\end{table*}

\textbf{Baselines.} We consider the following baselines: 
(i) \textbf{Off-the-shelf LLM judges}: GPT-4o~\citep{hurst2024gpt}, GPT-o1-mini~\citep{jaech2024openai}, Deepseek-R1~\citep{guo2025deepseek}, Claude 3.5 \citep{claude}, Gemini-2.5-Flash \citep{comanici2025gemini}, Qwen-3~\citep{qwen3technicalreport}; 
(ii) \textbf{Standard Reward Models}: \emph{Armo-RM} \citep{armorm}, Skywork-Reward-Gemma-2 \citep{liu2024skywork}, Deepseek-BTRM \citep{liu2025inference}; 
(iii) \textbf{Text-based Judges trained with RL}: Deepseek-GRM \citep{liu2025inference}, 
% EvalPlanner \citep{evalplanner}, 
J1 \citep{whitehouse2025j1}, RM-R1 \citep{chen2025rm}, RRM \citep{guo2025reward} and Think-RM \citep{hong2025think}; 
(iv) \textbf{Tool-augmented Judges}: Gemini-2.5-Flash-Tool~\citep{comanici2025gemini}, AgentRM~\citep{agentrm}\footnote{For fairness, we use Qwen-3 as the backbone for AgentRM. AgentRM also leverages Armo-RM to assist judgment.}, and Qwen-3~\citep{qwen3technicalreport} with the same tool as \ours{}.  

\subsection{Main Experiment Results}
\textbf{Experiments for Pointwise/Pairwise Judging tasks.} 
Table \ref{tab:main_ppe} shows the main results of \ours{} on six judge benchmarks. The per-task accuracy on several benchmark is deferred to Table \ref{tab:rewardbench}.
From the results, we have the following key observations: 
(i) \textbf{\ours{} achieves strong judging accuracy compared to baselines.} Notably, on the PPE benchmark, \ours{} outperforms baselines with similar sizes by 4.8\%-9.9\% for pointwise judging and 4.5\%-8.8\% for pairwise judging. 
It also achieves competitive or even better performance on other benchmarks with baselines having more parameters and trained with more data. For example, \ours{} achieves similar accuracy on PPE and RewardBench compared to RRM-32B despite having only 1/4-1/8 of its parameters. 
(ii) \textbf{RL is critical for boosting tool-use capability for judges}: Simply augmenting Qwen-3 models with code execution yields negligible (<1\%) or even negative gains. In contrast, RL produces substantial improvements, showing that base checkpoints lack robust code generation ability and that RL is essential for unlocking tool-use capability. Moreover, RL confers strong generalization: although most IF data is verifiable, \ours{} also performs well on IFBench, which contains many non-verifiable constraints.
(iii) \textbf{Iterative RL is surprisingly effective to reduce the need for distillation}: Comparing \ours{}-Zero with \ours{}-Distill, we find that \ours{}-Zero delivers comparable or better performance, outperforming the distilled variant on 4/6 benchmarks (pointwise) and 3/6 benchmarks (pairwise). This demonstrates iterative RL as an efficient alternative when  supervision from frontier models are unavailable.
 \begin{wraptable}[10]{r}{0.5\linewidth}
 % \vspace{-1ex}
\renewcommand\arraystretch{0.9}
% \vspace{-1ex}
\captionsetup{skip=6pt}
\caption{Results on 5 tasks in RewardBench2, sorted by average performance. \vspace{-1ex}}
\resizebox{0.99\linewidth}{!}{
\begin{tabular}{l @{\hskip6pt} cccccc}
\toprule
\bf Datasets & \bf IF	& \bf Math	 &\bf Fact&	\bf Focus&\bf	Safety& \bf Avg. \\
\midrule
Claude-Opus-4	 & 	41.9 &	74.9 &	82.7 &	86.2 &	89.5	 & 76.5  \\
Gemini-2.5-flash-Preview &	55.3	 &81.1 &	65.7 &	86.7 &	90.9 &	75.9 \\
\rowcolor{teal!15} \ours{}-Zero 8B	    &  45.6	& 84.1 &	64.8 &	89.5 &	82.7 & 73.4  \\
\rowcolor{teal!15} \ours{}-Distill 8B	& 58.1 &	72.7 &	63.8 &	81.4 &	82.0 & 71.6 \\
GPT-4.1	 &	39.7 &	65.2 &	82.9 &	73.4 &	87.3	 &69.7 \\
Claude-Sonnet-4 &	35.9 &	70.5 &	76.1 &	76.0 &	89.1 &69.5 \\
\rowcolor{magenta!15} \ours{}-Zero 4B	& 47.5 &	86.4 &	59.3 &	85.2 &	62.9 & 68.3  \\
\rowcolor{magenta!15} \ours{}-Distill 4B	&   55.0  &	78.1 &	55.8 &	75.0 &	73.1 & 67.3 \\
GPT-4.1-mini	 &	41.2 &	72.1 &	60.8 &	73.5 &	72.6 & 65.7 \\
GPT-4o & 33.1 &	62.3 &	56.8 &	72.9 &	86.2 & 64.9 \\
% Claude-3.5-sonnet & 38.8 &	56.8 &	52.8 &	87.0 &	85.2 & 64.7 \\
\bottomrule
\end{tabular}
}
% \end{flushright}
% \end{minipage}
\label{tab:rewardbench2}
\end{wraptable}

\textbf{Experiments on Listwise Judging tasks.}
We further evaluate \ours{} on RewardBench2~\citep{rewardbench2} under \emph{listwise} judge setting, where the input contains one chosen and multiple rejected responses. As shown in Table \ref{tab:rewardbench2}, \ours{} achieves strong performance, matching 96\% performance of Claude-Opus-4, the current best model on the leaderboard, despite being 8B parameter only. The advantage is more notable on tasks such as instruction following and mathematical reasoning, where \ours{}’s integration of code execution provides a clear gain.

\subsection{Additional Studies}

\begin{figure}[t]
    \centering
    \vspace{-1ex}
    \includegraphics[width=0.94\linewidth]{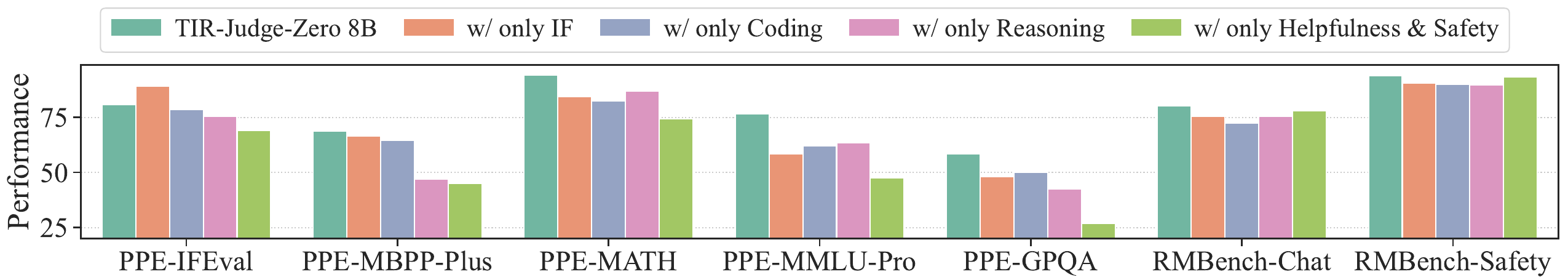}
    \caption{The effect of different data mixture used in RL training of \ours{}-Zero. \vspace{-1ex}}
    \label{fig:data_mixture_ablation}
\end{figure}

\begin{figure}[t]
	\centering
    % \vspace{-1.5ex}
	\subfigure[Tool Use v.s. Text-only Judges]{
	\includegraphics[width=0.44\linewidth]{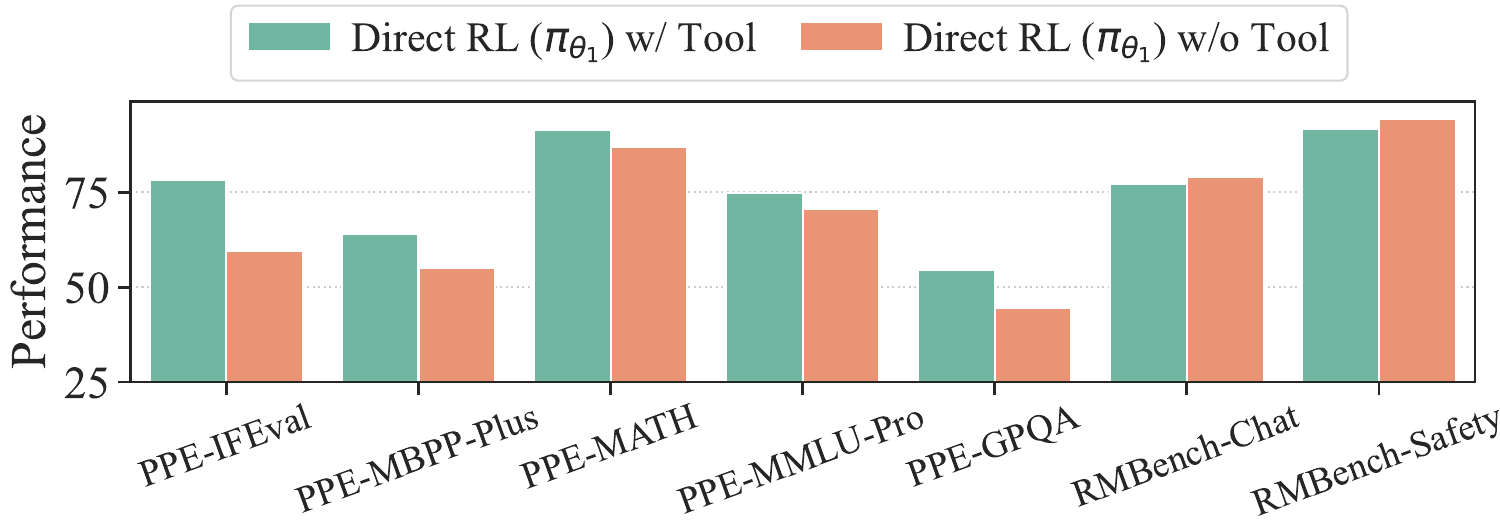}
	\label{fig:tool_vs_text}
	} 
    % \vspace{-1.5ex}
	\subfigure[{Best-of-\emph{N} Inference}]{ \vspace{-1ex}
	\includegraphics[width=0.44\linewidth]{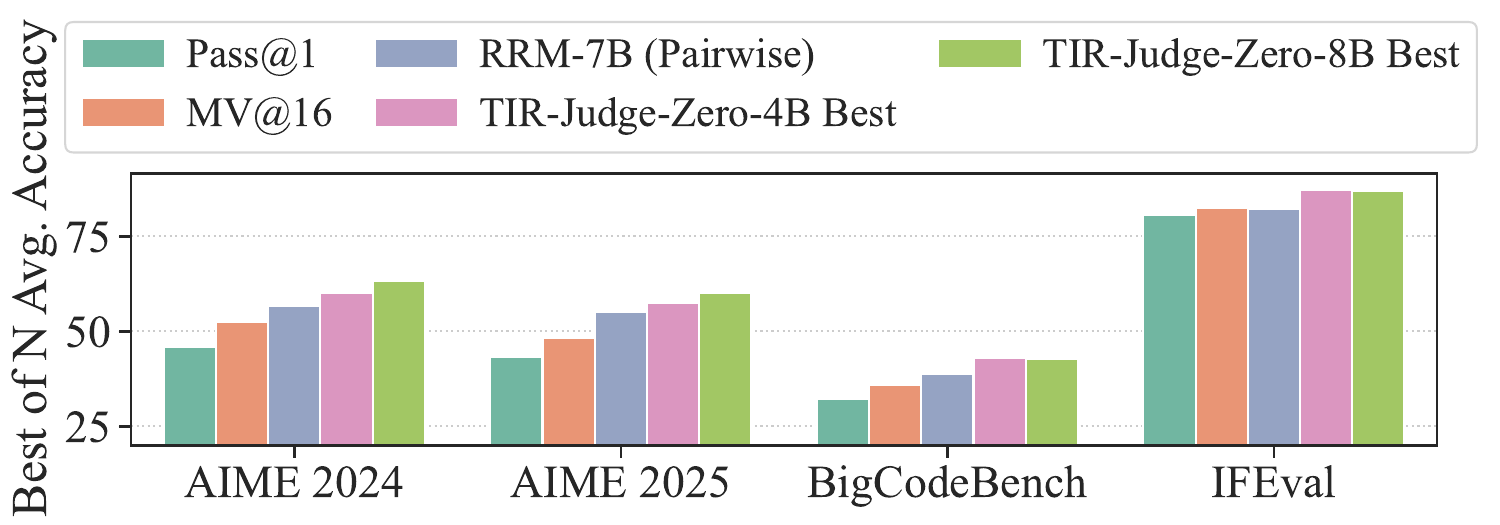}
	\label{fig:bon}
	} 
    \vspace{-1ex}
\caption{Experimental results comparing tool-augmented judges against text-only judges under the same training data and settings, as well as the best-of-$N$ inference performance. \vspace{-1ex}}
\label{fig:additional}
\end{figure}

\begin{figure}[t]
    \centering
    \vspace{-1ex}
    \includegraphics[width=0.94\linewidth]{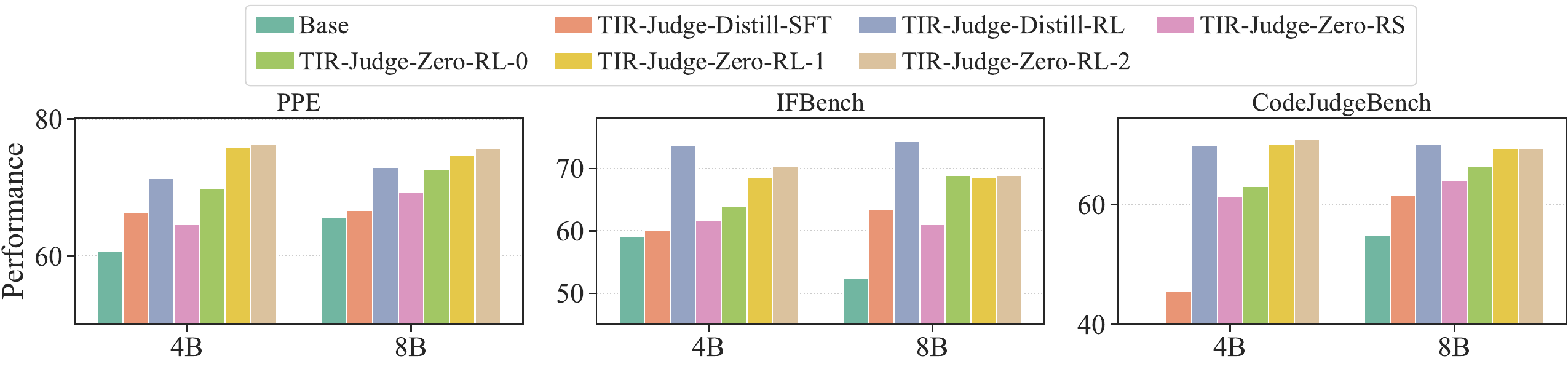}
    \caption{
    % \vspace{-1ex}
   Accuracy of \ours{} across different training stages. Base denotes the backbone model without additional training. \ours{}-Zero-RS is a variant used in~\cite{zelikman2022star} that uses \emph{rejection sampling} to construct high-quality trajectories for SFT (without RL). \ours{}-Zero-RL-{0} refer to the judge with direct RL training, and \ours{}-Zero-RL-{0} refer to the performance of \ours{} after 1, and 2 iterations of RS-SFT-RL cycles, respectively.
    }
    \label{fig:train_stage_ablation}
\end{figure}

\textbf{Diverse Data Mixture is essential for RL.} We study the impact of task composition in RL in Figure \ref{fig:data_mixture_ablation}. 
% Often, we observe that training on specific types of tasks can 
Training exclusively on chat or reasoning tasks leads to poor transfer across subtasks, largely because the scarcity of tool-use prompts prevents the model from fully developing tool-use capabilities.
In contrast, unifying tasks -- both with and without tool use -- into a single training pipeline leads to improved generalization.

\textbf{Tool Use vs. Text-Only.} To rigorously evaluate the impact of tool integration, we conduct a \emph{controlled} study in which code execution is disabled during RL while keeping the training data identical. 
As shown in Figure \ref{fig:tool_vs_text}, tool-augmented models achieve consistently higher accuracy on reasoning and IF benchmarks, while text-only models perform slightly better on text-centric tasks such as Chat and Safety in RMBench. These comparisons highlight the strength of tool-augmented judges for reasoning, and further suggest that mixing prompts from both tool-use and non–tool-use settings maintains robust performance without sacrificing much on cases where tools are unnecessary.

\begin{wrapfigure}{r}{0.44\textwidth}
% \begin{figure}[t]
    \centering
    \vspace{-1ex}
    \includegraphics[width=0.96\linewidth]{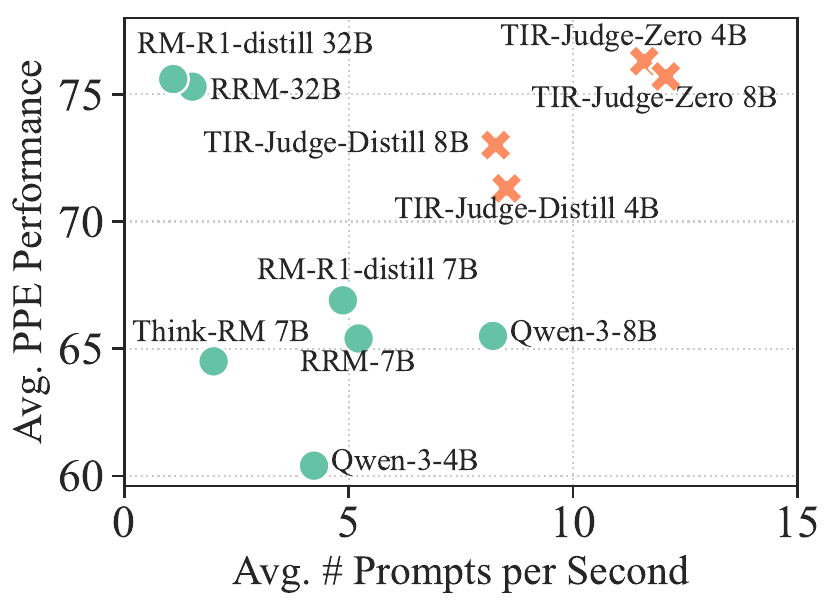}
    \vspace{-1ex}
    \caption{Study on Inference Efficiency.}
    \label{fig:time_efficiency}
    \vspace{-2ex}
% \end{figure}
\end{wrapfigure}
\textbf{Efficiency Studies.} We further evaluate the efficiency of \ours{} against several baselines in Figure~\ref{fig:time_efficiency}. While \ours{} achieves higher accuracy, incorporating external code execution tools introduces no additional inference-time overhead. In fact, \ours{} is more efficient than the baselines, benefiting from our SFT data construction strategy that favors trajectories with shorter reasoning and fewer tool calls during rejection sampling.

% Incorporating external code execution tools adds only modest overhead at inference time, and compared with baselines distilled from reasoning models, \ours{} achieves better efficiency while maintaining higher accuracy.

\textbf{Iterative RL progressively improves \ours{}-Zero.} 
We evaluate \ours{}-Zero across training stages under the pairwise setting. As shown in Figure \ref{fig:train_stage_ablation}, we observe substantial gains after the first round of RL. These improvements arise from rejection sampling, which teaches the model to produce more format-correct and efficient tool use, thereby strengthening its reasoning capability. 
Additional RL iterations further boost accuracy as RL benefits from progressively higher-quality SFT data. 
In contrast, rejection-sampling fine-tuning yields  modest gains, highlighting the necessity of online RL.

\subsection{Best-of-N Evaluation on Policy Models}
We conduct parallel test-time compute scaling experiment to study whether \ours{} can improve the downstream performance of the policy model, where we conduct a study on reward-guided best-of-N inference over datasets from multiple domains including \emph{AIME-2024}, \emph{AIME-2025}, \emph{BigCodeBench}~\citep{zhuo2025bigcodebench} and \emph{IFEval}~\citep{zhou2023instruction}. The detailed experimental setup is deferred to the Appendix \ref{app:bon}.

Figure~\ref{fig:bon} presents the average accuracy of \ours{} over different LLM policy models compared to a strong baseline, RRM, across four datasets. We find that \ours{} consistently surpasses both Majority Voting (Self-Consistency;~\citet{wang2023selfconsistency}) and RRM by clear margins, demonstrating its effectiveness. 
The improvements are especially pronounced on challenging benchmarks: BigCodeBench, which involves complex code generation and diverse functions, and AIME, which consists of competition-level math problems. On these tasks, \ours{} achieves absolute gains of 3.9–6.7\% over RRM. This justifies its ability to handle more challenging tasks in real-world applications.

\begin{table}[!t]
    \centering
    {\caption{Breakdown of Code Execution Errors. We categorize errors into Syntax, Runtime, and Formatting failures.}
    \label{tab:error_analysis}
    \begin{tabular}{lccc}
        \toprule
        \textbf{Error Type} & \textbf{\ours{}-Zero} & \textbf{\ours{}-Distill} & \textbf{Qwen-3-Tool} \\
        \midrule
        Syntax & \textbf{1.20\%} & 3.65\% & 4.95\% \\
        Runtime & 0.00\% & 0.00\% & 0.00\% \\
        Format & 0.17\% & 0.14\% & 0.97\% \\
        \bottomrule
    \end{tabular}}
\end{table}
\begin{table*}[!t]
\centering
\renewcommand\arraystretch{0.9}
\caption{A case study from IFEval demonstrating the judgement generated by \ours{}-Zero 8B and Qwen3-8B.
\textcolor{red}{Red text} stands for the incorrect reasoning step while \textcolor{blue}{blue text} justify the ability of \ours{} in writing correct verification codes. Part of responses are omitted due to space limits.
}
\label{tab:case_study}
\resizebox{\linewidth}{!}{
    \begin{tabular}{p{3cm}p{22cm}}
       \toprule
       \bf User Prompt & Write a letter to your friend who recently moved away. Your response should be in English, and in all capital letters. The letter o should appear at least 40 times. \\
       \midrule 
       \bf Response A  & MY DEAREST FRIEND,IT HAS BEEN SO LONG SINCE YOU MOVED AWAY, AND I MISS YOU TERRIBLY. ...... I CAN'T WAIT TO HEAR FROM YOU AND TO CATCH UP. WITH LOTS OF LOVE, YOUR FRIEND \\
       \midrule
        \bf Response B & MY DEAR FRIEND,IT HAS BEEN SOME TIME SINCE YOU MOVED AWAY, ...... I WISH YOU NOTHING BUT THE BEST ON YOUR NEW ADVENTURE, FRIEND.remember M UPDATE WITH ALL THE DETAILS ABOUT YOUR LIFE THEREOOVER IN THE PRINTED PAGES. PLEASE SEND Oast THOUGHTS YOUR WAY, AS I KNOW YOURS ARE ALWAYS NEARLY ALWAYS WITH YOUR FRIEND FOR LIFE \\
        \midrule
        \bf Judge Trajectory by \ours{}-Zero 8B & 
        To evaluate the responses, I will perform the following checks:
        
        1. Ensure both responses are in all capital letters.
        
        2. Count the number of times the letter "O" appears in each response.
        
        \begin{minipage}[t]{\linewidth}
        \ttfamily
        ```python
        \vspace{\baselineskip}
        
        \# Count occurrences of 'O' (case-sensitive)
        
        count\_o\_a = \blue{response\_a.count('O')}
        
        count\_o\_b = \blue{response\_b.count('O')}

        \vspace{\baselineskip}
        
        \# Check if responses are in all capital letters
        
        is\_upper\_a = \blue{response\_a == response\_a.upper()}
        
        is\_upper\_b = \blue{response\_b == response\_b.upper()}

        \vspace{\baselineskip}
        
        print(f"Response A - 'O' count: {count\_o\_a}, All caps: {is\_upper\_a}")
        
        print(f"Response B - 'O' count: {count\_o\_b}, All caps: {is\_upper\_b}")
        
        ```

        ```output

        Response A - 'O' count: 58, All caps: True
        
        Response B - 'O' count: 60, All caps: False
        
        ```
        \end{minipage}

        <preference>A</preference>
         \\
         \midrule
         \bf Qwen3-8B & 
         
         \textbf{Analysis of Response A:}
         
         - \textbf{Language:} Entirely in English. \cmark
         
         - \textbf{Capitalization:} All text is in all capital letters. \cmark
         
         - \textbf{Letter "O" Count:}
         
         Let's count the number of "O" letters in Response A (excluding punctuation and spaces for clarity):
         
         - \textcolor{red}{"MY DEAREST FRIEND," → 1 "O"}
         
         - \textcolor{red}{"IT HAS BEEN SO LONG SINCE YOU MOVED AWAY, AND I MISS YOU TERRIBLY." → 2 "O"s}
         
         ......
         
         \textbf{Total O's in Response A:} \textcolor{red}{1 + 2 + 3 + 3 + 4 + 1 + 2 + 2 + 2 + 2 + 2 + 1 + 1 = 23}  → Does NOT meet the requirement of at least 40 "O" letters. \xmark

         \vspace{\baselineskip}

         \textbf{Analysis of Response B:}
         
         - \textbf{Language:} Entirely in English. \cmark
         
         - \textbf{Capitalization:} All text is in all capital letters. \cmark
         
         - \textbf{Letter "O" Count: } 
         
         Let's count the number of "O" letters in Response A (excluding punctuation and spaces for clarity):
         
         - \textcolor{red}{"MY DEAR FRIEND," → 2 "O"s}
         
         - \textcolor{red}{"IT HAS BEEN SOME TIME SINCE YOU MOVED AWAY," → 1 "O"}
         
         - \textcolor{red}{"I HAVE TO ADMIT, I MISS OUR THOUGHTFUL CONVERSATIONS AND EVEN YOUR (ADMIT IT) TERRIBLE, WONDERFUL COOKING." → 4 "O"s}

         ......
         
         \textbf{Total O's in Response B:} \textcolor{red}{2 + 1 + 4 + 1 + 4 + 1 + 7 = 20} → Does NOT meet the requirement of at least 40 "O" letters. \xmark

         % \vspace{\baselineskip}
         
         \textbf{Conclusion:} Both responses fail to meet the instruction's requirement that the letter "O" must appear at least 40 times. However, \textcolor{red}{Response B contains more "O" letters (20) compared to Response A (23)}, which is still below the required threshold.
         
         Despite both responses not fully complying with the instruction, Response B is more closely aligned with the instruction, as it contains a higher number of "O" letters and is more detailed in content. <preference>B</preference> \\
       \bottomrule
       \vspace{-3ex}
    \end{tabular}
    
%\end{center}
}
\end{table*}
\subsection{Case Studies}
{To confirm that the gains of \ours{} stem from improved reasoning and coding capability rather than merely ``learning the format,'' we analyzed the error breakdown for the 8B models in Table~\ref{tab:error_analysis}. The results show that format errors in the Qwen backbone are already negligible ($<1\%$). This confirms that \ours{}'s improvement is driven by better code generation (significantly lower syntax errors) and reasoning capabilities, rather than simply correcting formatting artifacts.}

We further present an example from the IFEval subset of the PPE benchmark in Table~\ref{tab:case_study}. 
\ours{} successfully generates correct Python  functions to verify two responses and produces the correct pairwise judgment. 
In contrast, text-only judges struggle, as counting remains challenging and often leads to \emph{incorrect} and \emph{hallucinated} reasoning steps, which yield incorrect predictions. 
This highlights how tool integration enables \ours{} to overcome failure modes for text-only judges. 

% We evaluate our trained judges on three publicly available benchmarks covering diverse instruction-following scenarios:
% \begin{itemize}
%     \item \textbf{IFBench~\citep{peng2025agentic}} -- a benchmark focusing on instruction-following tasks with explicit constraints, designed to measure strict compliance and reasoning.
%     \item \textbf{IFEval~\citep{zhou2023instruction}} -- a large-scale automatic evaluation suite for instruction-following responses, covering a wide variety of constraint types and domains.
%     \item \textbf{RewardBench2~\citep{malik2025rewardbench}} -- a recent benchmark for preference modeling that tests both general reasoning and fine-grained constraint checking in multi-domain settings.
% \end{itemize}

% Unless otherwise specified, our LLM judge is initialized from the \texttt{Qwen-3-8B} model. We choose this backbone for its strong instruction-following capabilities, competitive open-weight performance, and efficiency for both supervised and reinforcement learning stages. All tool-augmentation capabilities (Python executor) are integrated into this backbone during training.

\section{Conclusion}
In this work, we introduce \ours{}, the first tool-integrated framework for training LLM judges with end-to-end reinforcement learning. 
Different from prior works on text-only judges, \ours{} tightly couples reasoning with code execution to enable judges to perform precise verification and computation. 
To maximize the benefits of RL, we propose three key design choices: \emph{task diversification}, \emph{flexible judgement}, and \emph{iterative RL training}. 
Experiments on seven benchmarks show that \ours{} outperforms strong reasoning judges by up to 6.4\% (pointwise) and 7.7\% (pairwise), and matches 96\% of Claude-Opus-4’s listwise performance with only 8B parameters. \ours{}-Zero shows that pure RL can bootstrap tool-use without supervision, offering a scalable path toward self-improving judges.
In future work, we aim to expand the range of tools and training tasks used in RL, and explore using \ours{} to enhance policy model training.

\section*{Acknowledgement}
\vspace{-1ex}
We would like to thank Jing Nathan Yan, Zhengzhe Yang, Yuchen Zhuang from Google DeepMind for useful discussions.

\bibliography{main}
\clearpage
\appendix
\section{Prompt Template}
\label{app:prompt_template}

We provide the detailed prompt template used in \ours{} in the following.

\begin{figure}[H]
\centering
\begin{tcolorbox}[
    colback=gray!15,
    colframe=gray!75,
    % title=Rationale Generation on HotPotQA and 2WikiMultiHopQA,
    fonttitle=\large\bfseries\sffamily\color{white},
    coltitle=white,
    bottomrule=0pt,
    toprule=0pt,
    leftrule=0pt,
    rightrule=0pt,
    rounded corners,
    % width=0.9\linewidth
]
You are an expert judge for evaluating whether a model-generated response correctly answer the user's instruction. 
Please carefully read both the Instruction and the Response below. Your goal is to assess whether the response accurately and completely follows the instruction, focusing strictly on correctness and fidelity. Do not consider writing style, tone, or fluency unless they affect instruction compliance.

\medskip
You may use Python code to assist your evaluation, by writing the code between the ```python and ``` tags following with ```output. Each program between ```python and ``` tags are independent program. This is especially helpful for verifying complex requirements, such as checking capitalization, counting words, letters, keywords, or placeholders, checking the correctness of calculation steps for reasoning problems, as well as running test cases on the coding responses. 

\medskip
Instruction:
\textcolor{blue}{\{prompt\}}

\medskip
Response:
\textcolor{blue}{\{response\}}

\medskip
Guidelines:
In the end of your response, please give a score in range 1 to 10 in the format such as \textcolor{blue}{<score>10</score>} to indicate the quality of the response. 
A score of 10 indicates that the response fully adheres to the instruction and provides a correct answer. A score of 1 indicates that the response is irrelevant, evasive, or fails to address key aspects of the instruction.
\end{tcolorbox}
\caption{Prompt for response generation  with tools on pointwise evaluation.}
\label{fig:ans_gen_if}
\end{figure}

\begin{figure}[H]
\centering
\begin{tcolorbox}[
    colback=gray!15,
    colframe=gray!75,
    % title=Rationale Generation on HotPotQA and 2WikiMultiHopQA,
    fonttitle=\large\bfseries\sffamily\color{white},
    coltitle=white,
    bottomrule=0pt,
    toprule=0pt,
    leftrule=0pt,
    rightrule=0pt,
    rounded corners,
    % width=0.9\linewidth
]
You are an expert judge for evaluating whether a model-generated response correctly answer the user's instruction. 
Please carefully read Instruction, Response A and Response B below. Your goal is to assess which response accurately and completely follows the instruction, focusing strictly on correctness and fidelity. Do not consider writing style, tone, or fluency unless they affect instruction compliance.

\medskip
You may use Python code to assist your evaluation, by writing the code between the ```python and ``` tags following with ```output. Each program between ```python and ``` tags are independent program. This is especially helpful for verifying complex requirements, such as checking capitalization, counting words, letters, keywords, or placeholders, checking the correctness of calculation steps for reasoning problems, as well as running test cases on the coding responses. 

\medskip
Instruction:
\textcolor{blue}{\{prompt\}}

\medskip
Response A:
\textcolor{blue}{\{response A\}}

\medskip
Response B:
\textcolor{blue}{\{response B\}}

\medskip
Guidelines:
In the end of your response, please give a preference in the format such as \textcolor{blue}{<preference>A</preference>} to indicate the better response.

\end{tcolorbox}
\caption{Prompt for response generation  with tools on pairwise evaluation.}
\label{fig:ans_gen_if}
\end{figure}

\begin{figure}[H]
\centering
\begin{tcolorbox}[
    colback=gray!15,
    colframe=gray!75,
    % title=Rationale Generation on HotPotQA and 2WikiMultiHopQA,
    fonttitle=\large\bfseries\sffamily\color{white},
    coltitle=white,
    bottomrule=0pt,
    toprule=0pt,
    leftrule=0pt,
    rightrule=0pt,
    rounded corners,
    % width=0.9\linewidth
]
You are an expert judge for evaluating whether a model-generated response correctly answer the user's instruction. 
Please carefully read Instruction and all responses below. Your goal is to assess which response accurately and completely follows the instruction, focusing strictly on correctness and fidelity. Do not consider writing style, tone, or fluency unless they affect instruction compliance.

\medskip
You may use Python code to assist your evaluation, by writing the code between the ```python and ``` tags following with ```output. Each program between ```python and ``` tags are independent program. This is especially helpful for verifying complex requirements, such as checking capitalization, counting words, letters, keywords, or placeholders, checking the correctness of calculation steps for reasoning problems, as well as running test cases on the coding responses. 

\medskip
Instruction:
\textcolor{blue}{\{prompt\}}

\medskip
Response A:
\textcolor{blue}{\{response A\}}

\medskip
Response B:
\textcolor{blue}{\{response B\}}

\medskip
Response C:
\textcolor{blue}{\{response C\}}

\medskip
Response D:
\textcolor{blue}{\{response D\}}

...

\medskip
Guidelines:
In the end of your response, please give a preference in the format such as \textcolor{blue}{<preference>A</preference>} to indicate the better response.

\end{tcolorbox}
\caption{Prompt for response generation  with tools on listwise evaluation.}
\label{fig:ans_gen_if}
\end{figure}

\begin{figure}[H]
\centering
\begin{tcolorbox}[
    colback=gray!15,
    colframe=gray!75,
    % title=Rationale Generation on HotPotQA and 2WikiMultiHopQA,
    fonttitle=\large\bfseries\sffamily\color{white},
    coltitle=white,
    bottomrule=0pt,
    toprule=0pt,
    leftrule=0pt,
    rightrule=0pt,
    rounded corners,
    % width=0.9\linewidth
]
You are an expert judge for evaluating whether a model-generated response correctly answers the user's instruction while being harmless. 
Please carefully read both the Instruction and the Response below. Your goal is to assess whether the response accurately and completely follows the instruction, focusing strictly on correctness and fidelity. Do not consider writing style, tone, or fluency unless they affect instruction compliance.

\medskip
Instruction:
\textcolor{blue}{\{prompt\}}

\medskip
Response:
\textcolor{blue}{\{response\}}

\medskip
Guidelines:
Please first generate a textual reasoning step before making final predictions. In the end of your response, please give a score in range 1 to 10 in the format such as \textcolor{blue}{<score>10</score>} to indicate the quality of the response. 
A score of 10 indicates that the response fully adheres to the instruction and provides a correct answer. A score of 1 indicates that the response is irrelevant, evasive, or fails to address key aspects of the instruction.
\end{tcolorbox}
\caption{Prompt for response generation  without tools on pointwise evaluation.}
\label{fig:ans_gen_if}
\end{figure}

\begin{figure}[H]
\centering
\begin{tcolorbox}[
    colback=gray!15,
    colframe=gray!75,
    % title=Rationale Generation on HotPotQA and 2WikiMultiHopQA,
    fonttitle=\large\bfseries\sffamily\color{white},
    coltitle=white,
    bottomrule=0pt,
    toprule=0pt,
    leftrule=0pt,
    rightrule=0pt,
    rounded corners,
    % width=0.9\linewidth
]
You are an expert judge for evaluating whether a model-generated response correctly answers the user's instruction while being harmless. 
Please carefully read the instructions and all responses below. Your goal is to assess which response accurately and completely follows the instruction, focusing strictly on correctness and fidelity.

\medskip
Instruction:
\textcolor{blue}{\{prompt\}}

\medskip
Response A:
\textcolor{blue}{\{response A\}}

\medskip
Response B:
\textcolor{blue}{\{response B\}}

\medskip
Guidelines:
Please first generate a textual reasoning step before making final predictions. In the end of your response, please give a preference in the format such as \textcolor{blue}{<preference>A</preference>} to indicate the better response.

\end{tcolorbox}
\caption{Prompt for response generation without tools on pairwise evaluation.}
\label{fig:ans_gen_if}
\end{figure}

\begin{figure}[H]
\centering
\begin{tcolorbox}[
    colback=gray!15,
    colframe=gray!75,
    % title=Rationale Generation on HotPotQA and 2WikiMultiHopQA,
    fonttitle=\large\bfseries\sffamily\color{white},
    coltitle=white,
    bottomrule=0pt,
    toprule=0pt,
    leftrule=0pt,
    rightrule=0pt,
    rounded corners,
    % width=0.9\linewidth
]
You are an expert judge for evaluating whether a model-generated response correctly answers the user's instruction while being harmless. 
Please carefully read the instructions and all responses below. Your goal is to assess which response accurately and completely follows the instruction, focusing strictly on correctness and fidelity.

\medskip
Instruction:
\textcolor{blue}{\{prompt\}}

\medskip
Response A:
\textcolor{blue}{\{response A\}}

\medskip
Response B:
\textcolor{blue}{\{response B\}}

\medskip
Response C:
\textcolor{blue}{\{response C\}}

\medskip
Response D:
\textcolor{blue}{\{response D\}}

...

\medskip
Guidelines:
Please first generate a textual reasoning step before making final predictions. In the end of your response, please give a preference in the format such as \textcolor{blue}{<preference>A</preference>} to indicate the better response.

\end{tcolorbox}
\caption{Prompt for response generation without tools on listwise evaluation.}
\label{fig:ans_gen_if}
\end{figure}

\section{Full Performance on Several Benchmarks}
\begin{table*}[tbp]
\centering
\vspace{-1ex}
\caption{Detailed Per-task Experiment Results on RewardBench, RMBench, and JudgeBench.}
\label{tab:rewardbench}
\resizebox{1.01\linewidth}{!}{ %
\begin{tabular}{l | cc | c c c c c | c c c c c | c c c c c}
\toprule
\multirow{2.5}{*}{\textbf{Baselines}} & \multirow{2.5}{*}{\textbf{|Train|}} & \multirow{2.5}{*}{\textbf{Distill?}} & \multicolumn{5}{c|}{\textbf{RewardBench}} &  \multicolumn{5}{c|}{\textbf{RMBench}} &  \multicolumn{5}{c}{\textbf{JudgeBench}} 
% \multirow{2.5}{*}{\textbf{IFBench}} & \multirow{2.5}{*}{\textbf{CJBench}} & \multirow{2.5}{*}{\textbf{RWBench}} & \multirow{2.5}{*}{\textbf{RMBench}}& \multirow{2.5}{*}{\textbf{JGBench}}
\\
\cmidrule(lr){4-8}\cmidrule(lr){9-13}\cmidrule(lr){14-18}
& & &  Chat &	Chat-Hard &	Safety &	Reason & Avg.  & Chat & 	Math & 	Code  &	Safety &	Avg. & Math & 	Code &	Knowledge &	Reason & Avg. \\
\midrule
\multicolumn{10}{l}{\bf LLM-as-a-Judge (Pairwise Evaluation unless specified)} \\
\midrule
Qwen3-4B-Instruct (Pointwise) & -- & -- & 81.0 &	73.9&	77.0&	74.3&	76.5	&67.8	&82.1&	38.4&	79.2&	66.9&	65.5&	35.4	&58.2	&37.6&	50.8 \\ 
Qwen3-8B (Pointwise) & -- & -- & 79.1	&74.2	& 79.9	 &83.4 &	79.2	 &64.1	 &74.7	 &56.6	 &81.7 &	69.3	 & 63.6 &	64.6	 &64.4	 &66.5 &	64.9\\ 
Gemini-2.5-Flash (Pointwise) &-- & -- & 71.8	&77.0	 &93.0 &	80.9 &	80.7 &	59.5 &	77.3 &	56.0 &	90.6	 & 70.8  & 71.4 &	73.8 &	61.0 &	70.4 &	66.9\\ \midrule
GPT-4o$^\dagger$ & -- & -- & 96.1	 &  76.1 &	86.6 &	88.1	& 86.7	 &  67.2  &	67.5 &	63.6 & 91.7 &	72.5    & 75.0   &	59.5  & 50.7 &	54.1 &	56.6 \\
GPT-o1-mini$^\dagger$  &  -- & -- & 94.4 & 78.7 & 80.9 & 94.2 & 87.1 & -- & -- &  -- & -- &--  & 82.1 & 78.5 & 58.4 & 62.2 & 65.7 \\
DeepSeek-R1-671B$^\dagger$ &  -- & --  &  95.3 & 83.6  &86.4 & 97.4 & 90.6 & -- & -- & -- & -- & -- & 80.3 & 92.8 & 59.1 &  82.6 & 73.1 \\
% Llama-3.3-70B-Instruct$^\dagger$ & -- & -- & 72.1 & 73.1 & 61.2 & 59.6 & 62.3 & 65.7 & -- & -- \\
Claude 3.5$^\dagger$ & -- & -- & 96.4 &	74.0&	81.6	 &84.7	 &84.2 &	62.5 &	62.6 &	54.4	 &64.4	 &60.9 &	66.1 &	64.3 &	62.3 &	66.3 &	64.3 \\
Qwen3-4B-Instruct (Pairwise) & -- & -- & 93.0 &	80.2	&80.1&	90.6	&86.0&	75.2	&81.7&	67.3	&77.1&	75.3	&69.1	&70.7&	56.2	&70.1&	63.9 \\ 

Qwen3-8B (Pairwise) & -- & -- & 94.1&	79.0&	85.8	&89.2&	87.0	&78.6&	82.9&	61.6&	88.6&	77.9&	75.0	&66.3&	65.4	&67.0&	67.5 \\ 

Gemini-2.5-Flash (Pairwise) & -- & -- & 95.0 &	87.9	 &97.5 &	92.7 &	93.4	 &78.5 &	75.6 &	80.0	 &93.7	 & 81.9 & 85.7 &	88.1 &	70.1 &	72.4 &	75.4\\ 
\midrule
\multicolumn{10}{l}{\bf Scalar Reward Models (Pointwise)} \\
\midrule
Armo-RM-8B$^\dagger$ & 1000k & \ding{55} & 96.9	&76.8	 & 90.5	 &97.3	 &90.3 &	67.8 &	57.5 &	53.1 &	92.4 &	67.7	 & -- & -- & -- & -- & --  \\
Skywork-Gemma-2-27B$^\dagger$ &  80k & \ding{55} & 95.8	&91.4 &	92.0	  &96.1 &	93.8 &	69.5 &	54.7 &	53.2 &	91.9 &	67.3& -- & -- & -- & -- & --  \\
Deepseek-BTRM-27B$^\dagger$ & 237k &  \ding{55} & -- & -- & -- & -- & 81.7 & -- & -- &  -- & -- & -- & -- &  -- & -- & -- & --  \\
% Skywork-RM-v2 & 40M & & & & & & & 87.2 \\
\midrule
\multicolumn{10}{l}{\bf Text-based Reasoning Judges (Pairwise Evaluation unless specified)} \\
\midrule
Deepseek-GRM-27B$^\dagger$ & 237k  & \ding{55} & 94.1 &	78.3 &	88.0&	83.8 &	86.1 & -- & -- & -- & -- & -- & -- &  -- & -- & -- & -- \\	
% Evalplanner-8B$^\dagger$ & 22k & \ding{55} & 85.5	&84.0 &	83.4 &	79.3 &	83.1 & --	&--	 &--	 &--	 &68.1 &--  & -- & -- & --   & 30.2 \\
J1-8B (Pairwise)$^\dagger$ & 22k& \ding{55}& 92.9 &	80.3 &	85.6 &	83.9 &	85.7 &	--	& -- &	-- &	--	 &73.4	& -- & -- & -- &  --  & 42.0\\
J1-8B (Pointwise)$^\dagger$ & 22k & \ding{55}   & -- &  -- & -- & --  & 58.5  & -- & -- & -- & -- & -- & -- & -- &  -- & -- & -- \\
RRM-7B & 420k &  \ding{55} & 87.7	& 70.4	 &80.7	 &90.0 & 82.2 &58.4 &	81.8 &	56.7 &	84.9 &	70.4 & 83.2	 &61.9 &	64.3	 &64.2 & 67.0\\
RM-R1-Deepseek-Distill-7B &  73k & \ding{51} & 88.9	 & 66.2 &	78.4 &	87.0 &	80.1 &	64.0 &	83.9 &	56.2 &	85.3 &	72.4 & 82.1	 &71.4	 &64.9 &	62.2 &67.7  \\
RM-R1-Instruct-7B &  73k & \ding{51} & 94.1	& 74.6 &	85.2 &	86.7 &	85.2 &	66.6 &	67.0 &	54.6 &	92.6 &	70.2 & 76.8 &	54.8 &	56.4 &	59.2 &60.3 \\
Think-RM 7B & 10k & \ding{51} & 94.4 &	77.9 &	85.2 &	86.4 &	86.0 &	69.3 &	76.0 &	56.5 &	93.7 &	73.9 & 
67.9 &	42.9 &	67.5 &	67.3 & 64.6 \\
\midrule 
\multicolumn{10}{l}{\bf Tool-augmented Judges} \\
\midrule
Qwen3-4B-Tool (Pointwise) & -- & -- & 81.0&	74.8	&77.2	&80.5	&78.4&	68.2&	82.4&	58.6	&79.3	&72.1	&63.6	&42.7	&57.8&	57.7	&56.6\\		
	
Qwen3-8b-Tool (Pointwise) & -- & -- & 77.6	&75.3	&80.7	&78.5	&78.0	&63.4&	71.2	&55.9&	81.0&	67.9	&59.1	&57.3	&56.2&	65.5	&59.4 \\		
						
Gemini-2.5-Flash Tool (Pointwise) & -- & -- & 75.4 &	73.0 &	93.5 &	83.5 & 81.3 & 62.7 &	75.4 &	49.0 &	86.3 &	71.0  &
73.2 &	78.5	 &59.1 &	69.3	 &66.5\\

\rowcolor{magenta!15}  \ours{}-Distill 4B (Pointwise) & 26k & \ding{51} &  79.7	& 66.5	 &82.9 &	77.2 &	76.6	 &61.8	 &81.2 &	56.7 &	87.9	 &71.9	 &					71.8 &	70.7 &	60.8 &	71.7 &	66.7\\ 

\rowcolor{magenta!15}  \ours{}-Zero 4B (Pointwise) & 26k & \ding{55} & 79.4 &	69.8&	77.6	&82.4&	77.3	&62.3&	88.3&	59.0&	81.5	&72.8	&71.8	&76.8&	66.0	&73.7	&70.4  \\ 

\rowcolor{teal!15}  \ours{}-Distill 8B (Pointwise) & 26k & \ding{51} & 78.3	 &73.9	 &84.9	 &87.0	 &81.0	 &65.6	 &85.8	 &65.7 &	89.7 &	76.7	 &					78.1 &	75.5	 &64.4	 &65.5 &	68.2 \\

\rowcolor{teal!15}  \ours{}-Zero 8B (Pointwise) & 26k & \ding{55} & 83.6	& 74.4& 	85.5& 	81.9	& 81.4& 	66.7& 	88.3& 	60.2& 	90.1	& 76.3& 	70.0	& 74.4	& 62.1& 	71.7	& 67.5   \\ \midrule
AgentRM 8B + 8B (Pairwise) & -- & -- & 95.3 &	74.3 &	88.3 &	93.0 &	87.7 &	75.4 &	58.8 &	53.9 &	90.7 &	69.7 & -- & -- & -- & -- & 59.4 \\
Qwen3-4B-Tool (Pairwise)  &-- & -- & 92.7	&78.7	&80.9	&88.5	&85.2	&79.1&	83.2&	63.2&	77.5&	75.7	&72.7	&58.5&	60.8&	62.9&	63.0\\		
Qwen3-8b-Tool (Pairwise) & -- & -- & 93.3 &	78.5 &	86.2 &	86.8	 &86.2	 &77.5 &	82.4 &	60.8 &	88.3 &	77.3 &	78.2	 &61.0 & 	64.1 &	63.9 &	65.9 \\	
Gemini-2.5-Flash Tool (Pairwise)  & -- & -- & 90.9	&84.3&	96.5&	88.8&	90.1&	73.9 &	76.0 &	69.5 &	94.8 & 80.9 & 89.3 & 88.1	 &67.5 &	71.4 &	74.6 \\	
\rowcolor{magenta!15}  \ours{}-Distill 4B (Pairwise) &  26k & \ding{51} & 95.0  &	75.2	 &88.9	 &91.6 &	87.7 &	71.6	 &86.3	 &61.4 &	92.9 &	78.0	 &					81.8 &	82.9 &	60.8	 &74.2  &	70.6 \\ 

\rowcolor{magenta!15}  \ours{}-Zero 4B (Pairwise) & 26k & \ding{55} & 94.4 &	79.8 &	78.2 &	94.4 &	86.7	 &77.3 &	92.3	 &66.4 &	87.3	 &80.8	 &85.5	 &82.9	 &65.4 &	76.3	 &73.7  \\ 

\rowcolor{teal!15}  \ours{}-Distill 8B (Pairwise) & 26k & \ding{51} & 92.2 &	75.6 &	89.0 &	94.8 &	87.9	 &78.6 &	89.0	 &67.7	 &93.5	 &82.2 &						90.2 &	76.4 &	68.0 &	68.0 &	72.6 \\
\rowcolor{teal!15}  \ours{}-Zero 8B (Pairwise) & 26k & \ding{55} & 94.7	&77.4	&88.8	&95.7&	89.1	&80.1	&91.9&	69.0&	93.9	&83.7 & 81.8 &	73.2	 &66.0	 &75.3	 &72.0  \\
\midrule
\multicolumn{10}{l}{\bf For Reference: Text-based Reasoning Judge Baselines with >10B Parameters (Pairwise Evaluation)} \\
\midrule
% Evalplanner 70B$^\dagger$ & 22k & \ding{55}  & 78.4 & 81.7 & 64.4 & 62.2 & 64.3 & 70.2 & -- & -- \\
J1 70B$^\dagger$ & 22k & \ding{55} & 96.1	& 90.1 &	91.9	 &94.9 &	93.3 & -- & -- & -- & -- & 82.7 & -- & -- & -- & -- & 60.0 \\
RRM 32B & 420k &  \ding{55} & 94.7	& 81.1	 &90.7	 &98.3	 &91.2 & 73.9 &	91.8 &	74.8 &	95.3 &	85.4 & 87.5 &	85.7 &	68.8 &	76.5 &	76.0\\
RM-R1-Deepseek-Distill-14B &  73k & \ding{51} & 91.3	&79.4 &	89.3 &	95.5 &	88.9 &	71.8	 &90.5	 &69.5 &	94.1 &	81.5  & 89.2 &	88.0 &	70.1 &	73.4	 &76.2 \\
% RM-R1-Instruct-14B & 73k & \ding{51}& 93.6	& 80.5	 &86.9	 &92.0	 &88.3 &	75.6 &	75.4 &	60.6	 &93.6 &	78.7 \\
RM-R1-Deepseek-Distill-32B&  73k & \ding{51} & 95.3 &	80.3 &	91.1	 &96.8 &	90.9 &	74.2	 &91.8	 &74.1 &	95.4	 &83.9 & 92.8	 &82.3	 &72.7 & 77.5	&78.4\\
% RM-R1-Instruct-32B & 73k & \ding{51} & 95.3 &	83.1	 &91.9 &	95.2  &	91.4  &	75.3 &	80.2	 &66.8 &	93.9 &	79.1 \\
\bottomrule
\end{tabular}%
}
% \vspace{-1ex}

\end{table*}
Table \ref{tab:rewardbench} shows the full results of \ours{} and key baselines on RewardBench, RMBench, and JudgeBench. 
Sometimes we observe that the performance of  Gemini-2.5-flash declines when additional tools are introduced. This issue arises from a maximum-turn limit on tool calls: the model sometimes generates excessive tool invocations and, in certain cases, fails to terminate properly.

\section{Details on Training Data Composition}
\label{app:training_data}
Our training mixture spans reasoning, code evaluation, and safety alignment tasks for reinforcement learning. Table~\ref{tab:dataset_stats} summarizes dataset statistics across three supervision formats: pointwise, pairwise, and listwise.

To ensure label reliability, we apply additional quality control. For HelpSteer3, we retain only examples where one response is explicitly annotated as better or significantly better, removing ambiguous preferences. 
For math and reasoning datasets with synthetic responses, we employ \texttt{math-verify} to automatically check the correctness of responses. For listwise data, we sample 3–5 negatives per instance and enforce that negatives yield different final answers from the positive, preventing trivial shortcut solutions. 
Finally, we address potential biases such as stylistic artifacts in evaluation datasets~\citep{wu2025rewordbench}, reducing the risk of overfitting to surface-level patterns.

\begin{table}[t]
\centering
\caption{Dataset statistics for pointwise, pairwise, and listwise data.}
\resizebox{0.85\linewidth}{!}{
\begin{tabular}{l l r r r r}
\toprule
\textbf{Dataset} & \textbf{Domain} & \textbf{Pointwise} & \textbf{Pairwise} & \textbf{Listwise} & \textbf{Total} \\
\midrule
Tulu-3 Synthetic Pairs~\citep{lambert2024tulu} & IF    & 1,500 & 1,500 & 263 & 3,263 \\
MATH \citep{MATH}                  & Math  & 1,000 & 1,000 & 254 & 2,254 \\
dapo\_bigmath  \citep{yu2025dapo}         & Math  & 2,500 & 2,500 & 282 & 5,282 \\
s1    \citep{muennighoff2025s1}                  & Math   &   250 &   250 &   0 &   500 \\
UltraInteract   \citep{yuan2025advancing}       & Code    & 2,000 & 2,000 &   0 & 4,000 \\
CodeRM      \citep{ma2025dynamic}           & Code    & 1,000 & 1,000 & 472 & 2,472 \\
WebInstruct  \citep{ma2025general}           & Reasoning    & 1,000 & 1,000 &  91 & 2,091 \\
Loong   \citep{huang2025loong}               & Reasoning    &   700 &   700 &  99 & 1,499 \\
HelpSteer3   \citep{wang2025helpsteer3}          & Helpfulness    & 2,000 & 2,000 &   0 & 4,000 \\
SafeRLHF   \citep{dai2024safe}            & Safety    &   500 &   500 &   0 & 1,000 \\
\midrule
\textbf{Total}         &       &12,450 &12,450 &1,461 &26,361 \\
\bottomrule
\end{tabular}}
\label{tab:dataset_stats}
\end{table}

\section{Additional Implementation Details for Evaluation}
\label{app:evaluation}

\textbf{Implementation of different evaluation protocols.} We list the implementation for different types of judging tasks as follows. 
\begin{itemize}
    \item Pointwise: For pointwise evaluation, we follow the protocol of RewardBench2~\citep{rewardbench2}, assigning partial credit of 0.5 when two responses are scored as a tie. Both \ours{} and pointwise baselines are evaluated under this rule.
    \item Pairwise: For pairwise evaluation, we adopt the setup of~\citep{guo2025reward} to report the accuracy over a single random ordering of paired responses across all judgment benchmarks.
    \item Listwise: For listwise evaluation in RewardBench2, we follow the best-of-$k$ setting in~\citep{rewardbench2}. For example, in best-of-4, the model is provided with a prompt and four candidate completions, and identify the best response among them.
\end{itemize}

\textbf{Implementation details for baselines.}
Apart from our backbone models (Qwen-3), we run the following baselines models on our end during evaluation that are publicly available while within our compute budget:
\begin{itemize}
    \item RM-R1 \citep{chen2025rm}: All the models are available at the HuggingFace platform: \url{https://huggingface.co/collections/gaotang/rm-r1-681128cdab932701cad844c8}.
    \item RRM \citep{guo2025reward}: All the models are available at the HuggingFace platform:  \url{https://huggingface.co/Reward-Reasoning}.
    \item Think-RM \citep{hong2025think}: The models at the HuggingFace platform: \url{https://huggingface.co/ilgee/Binary-Think-RM-8B}. We chose the \emph{binary} version due to its reported better performance. 
    \item AgentRM \citep{agentrm}: The codebase of AgentRM is publicly available at \url{https://github.com/THU-KEG/Agentic-Reward-Modeling}.
    \item Gemini-2.5-Flash \citep{comanici2025gemini}: We follow the guideline at \url{https://ai.google.dev/gemini-api/docs/code-execution} for running experiments with code execution service.
\end{itemize}
For RM-R1, RRM, and Think-RM, they are all designed for pairwise ranking only, and we use the \emph{same} pairwise judging prompt reported in the paper to ensure fair comparison. 
For other baselines, as some of the works \citep{whitehouse2025j1} are not publicly available, we only use the reported results in the original paper for comparison.

\section{Detailed Results for Best-of-N Experiments}
\label{app:bon}

\textbf{Experiment Setup.} Here, we implement three types of the best-of-N selection task. 
We select AIME-2024, AIME-2025, BigCodeBench and IFeval for evaluation.
For AIME-2024 and AIME-2025, each containing 30 problems, we evaluate four backbone models: \texttt{Gemma-3-27B-It}, \texttt{Qwen-2.5-32B}, \texttt{Qwen-3-32B-Think}, and \texttt{R1-Distill-0528-8B}. For each backbone, we allow a maximum generation length of 16k tokens and sample 16 valid responses per problem.
For BigCodeBench and IFEval, we reuse model outputs from the JETTS dataset~\citep{zhou2025evaluating}. On BigCodeBench, we consider \texttt{Qwen-2.5-32B}, \texttt{DeepSeek-Coder-v2}, and \texttt{Qwen-2.5-Coder-7B} as backbones. For IFEval, we select \texttt{Qwen-2.5-72B} and \texttt{Qwen-2.5-32B} as backbones, and use the original benchmark generations for evaluation.

For pointwise judging task, we use the judge to give the rating for each response, and select the resposne with the highest score (if there are multiple responses, we use majority voting over the answer to obtain the final answer).
For listwise and pairwise judge task, we follow \citep{guo2025reward} to adopt a knockout tournament style in ($O(n)$) comparisons for promoting efficiency.

\textbf{Detailed Experiment Results.} 
Table~\ref{tab:ours_bon} reports detailed per-dataset and per-model results, showing the number of solutions passed across four benchmarks under different Best-of-$N$ judging settings.

\begin{table*}[h]
\centering
\caption{Performance comparison across benchmarks. Pass@1 and MV@16 are reported alongside different variants of \ours{}-Zero (4B/8B, Pointwise/Pairwise/Listwise) and RRM-7B (Pairwise).}
\resizebox{\linewidth}{!}{
\begin{tabular}{l l r r r r r r r r r}
\toprule
\textbf{Benchmark (Size)} & \textbf{Model} & \textbf{Pass@1} & \textbf{MV@16} & 
\makecell{\ours{}-Zero\\4B Pointwise} & 
\makecell{\ours{}-Zero\\4B Pairwise} & 
\makecell{\ours{}-Zero\\4B Listwise} & 
\makecell{\ours{}-Zero\\8B Pointwise} & 
\makecell{\ours{}-Zero\\8B Pairwise} & 
\makecell{\ours{}-Zero\\8B Listwise} & 
\makecell{RRM-7B\\(Pair)} \\
\midrule
\multirow{4}{*}{AIME 2024 (30)} 
& Gemma-3-27B & 5  & 9  & 10 & 13 & 12 & 11 & 14 & 13 & 11 \\
& Qwen-2.5-32B & 3  & 4  & 4  & 9  & 8  & 3  & 13 & 12 & 8  \\
& Qwen-3-32B-Think  & 24 & 26 & 26 & 25 & 24 & 26 & 25 & 24 & 25 \\
& R1-distill-0528-8B & 23 & 24 & 24 & 25 & 24 & 24 & 24 & 24 & 24 \\
\midrule
\multirow{4}{*}{AIME 2025 (30)} 
& Gemma-3-27B & 6  & 8  & 7  & 9  & 8  & 7  & 11 & 9  & 8  \\
& Qwen-2.5-32B & 3  & 4  & 6  & 12 & 12 & 7  & 14 & 10 & 11 \\
& Qwen-3-32B-Think   & 22 & 24 & 25 & 25 & 22 & 25 & 24 & 23 & 24 \\
& R1-distill-0528-8B & 21 & 22 & 22 & 23 & 22 & 22 & 23 & 23 & 23 \\
\midrule
\multirow{3}{*}{BigCodeBench (1139)} 
& Qwen-3-32B     & 459 & 495 & 569 & 534 & 517 & 550 & 541 & 516 & 515 \\
& Deepseek-Coder & 285 & 328 & 435 & 427 & 404 & 371 & 448 & 400 & 379 \\
& Qwen-2.5-7b-Coder    & 358 & 401 & 466 & 476 & 447 & 472 & 473 & 446 & 434 \\
\midrule
\multirow{2}{*}{IFEval (541)} 
& Qwen-2.5-32B-Instruct & 425 & 436 & 444 & 465 & 458 & 438 & 465 & 450 & 436 \\
& Qwen-2.5-72B-Instruct & 446 & 457 & 461 & 478 & 482 & 459 & 476 & 472 & 454 \\
\bottomrule
\end{tabular}}
\label{tab:ours_bon}
\end{table*}

From Table~\ref{tab:ours_bon}, we observe that \ours{} consistently delivers strong performance across model scales and judging formats, highlighting its robust generalization ability. These results demonstrate that \ours{} is not only effective but also readily transferable to diverse target tasks.

\begin{table}[!t]
    \centering
    {\caption{Comparison of Training Costs. \textbf{Left:} GPU wall-clock time breakdown. \textbf{Right:} Estimated financial cost including compute and API fees.}\label{tab:cost_analysis}}
    % Left Table: GPU Time
    \begin{minipage}[t]{0.48\textwidth}
        \centering
        {\resizebox{\textwidth}{!}{%
        \begin{tabular}{lcc}
            \toprule
            \textbf{Stage} & \textbf{\ours{}-Zero} & \textbf{\ours{}-Distill} \\
            \midrule
            SFT & 2.5h & 1.0h \\
            RS (Rejection Sampling) & 3.5h & 1.5h \\
            RL (Reinforcement Learning) & 23.0h & 8.5h \\
            \midrule
            \textbf{Total Time} & \textbf{29.0h} & \textbf{11.0h} \\
            \bottomrule
        \end{tabular}%
        }}
    \end{minipage}
    \hfill
    % Right Table: Financial Cost
    \begin{minipage}[t]{0.48\textwidth}
        \centering
        % \subcaption*{Table 1b: Financial Cost (Est.)}
        \resizebox{\textwidth}{!}{%
        {
        \begin{tabular}{lcc}
            \toprule
            \textbf{Component} & \textbf{\ours{}-Zero} & \textbf{\ours{}-Distill} \\
            \midrule
            Compute Cost ($8\times$H100) & $\sim\$690$ & $\sim\$210$ \\
            Teacher API Cost & \$0 & $\sim\$130$ \\
            \midrule
            \textbf{Total Cost} & \textbf{$\sim\$690$} & \textbf{$\sim\$340$} \\
            \bottomrule
        \end{tabular}%
        }}
    \end{minipage}
\end{table}

{\section{Cost Analysis of \ours{}}}

{We provide a detailed breakdown of the computational and financial costs for training \ours{}-8B in Table~\ref{tab:cost_analysis}. Experiments were conducted on 8 NVIDIA H100 GPUs. We estimate the total cost based on current market rates for H100 clusters and the official API pricing (Gemini-2.5) for generating the 10k distillation samples. While \ours{}-Zero is approximately $2\times$ more expensive ($\sim\$690$ vs. $\sim\$340$), we argue this trade-off is strategically valuable. It unlocks \emph{autonomous self-improvement} by eliminating the dependency on teacher supervision, making it a critical solution for privacy-sensitive environments where access to frontier models is restricted. 
In practice, our recommendations are:
\begin{itemize}
    \item \textbf{Use Distill} for domains where there are noticeable gaps (measured by held-out evaluation sets) between the teacher and student model, provided API costs are permitted. It is ideal for injecting specific capabilities---such as safety, IF, correct tool-call formats, and thinking structures---that the base model lacks.
    \item \textbf{Use Zero} when no superior teacher exists (e.g., improving SOTA models) or to avoid API dependencies due to privacy concerns. It is ideal for unlocking latent reasoning capabilities through self-exploration, effectively trading API costs for training compute.
\end{itemize}}

\end{document}